\documentclass[sigconf]{acmart}
\makeatletter
\let\@authorsaddresses\@empty
\makeatother

\renewcommand\footnotetextcopyrightpermission[1]{} % removes footnote with conference information in first column
\usepackage{times}
\usepackage{epsfig}
\usepackage{graphicx}
\usepackage{amsmath}
\usepackage{booktabs}
\usepackage{makecell}
\usepackage{subcaption}
\usepackage{bm}
\usepackage{algorithm}
\usepackage{algpseudocode}
\usepackage{mathtools}
\usepackage{multirow}
\usepackage{multicol}
\usepackage{comment}
\usepackage{tabularx}
\newcolumntype{Y}{>{\centering\arraybackslash}X}

\usepackage{xcolor, colortbl}
\usepackage{wrapfig}
\definecolor{grey}{rgb}{0.9, 0.9, 0.9}

\definecolor{grn}{rgb}{0.1, 0.6, 0.1}
\definecolor{mgt}{rgb}{0.6, 0.1, 0.6}
\usepackage{array}

\begin{document}

%%
%% The "title" command has an optional parameter,
%% allowing the author to define a "short title" to be used in page headers.
\title{DiffGAR: Model-Agnostic Restoration from Generative Artifacts Using Image-to-Image Diffusion Models}

\author{Yueqin Yin}
\affiliation{%
  \institution{School of Artificial Intelligence, University of Chinese Academy of Sciences \\ Institute of Automation, Chinese Academy of Sciences}
  \city{Beijing}
  \country{China}
  }
  
\author{Lianghua Huang}
\affiliation{%
  \institution{Machine Intelligence Technology Lab, Alibaba Group}
  \city{Beijing}
  \country{China}
}
  
\author{Yu Liu}
\affiliation{%
  \institution{Machine Intelligence Technology Lab, Alibaba Group}
  \city{Beijing}
  \country{China}
 } 

\author{Kaiqi Huang}
\affiliation{%
  \institution{School of Artificial Intelligence, University of Chinese Academy of Sciences \\ Institute of Automation, Chinese Academy of Sciences \\ CAS Center for Excellence in Brain Science and Intelligence Technology, Beijing, China}
  \city{}
  \country{}}

\begin{abstract}
  Recent generative models show impressive results in photo-realistic image generation. However, artifacts often inevitably appear in the generated results, leading to downgraded user experience and reduced performance in downstream tasks.
This work aims to develop a plugin post-processing module for diverse generative models, which can faithfully restore images from diverse generative artifacts. 
This is challenging because: (1) Unlike traditional degradation patterns, generative artifacts are non-linear and the transformation function is highly complex. (2) There are no readily available \textit{artifact-image} pairs. (3) Different from model-specific anti-artifact methods, a model-agnostic framework views the generator as a black-box machine and has no access to the architecture details. 
In this work, we first design a group of mechanisms to simulate generative artifacts of popular generators (i.e., GANs, autoregressive models, and diffusion models), given real images.
Second, we implement the model-agnostic anti-artifact framework as an image-to-image diffusion model, due to its advantage in generation quality and capacity. 
Finally, we design a conditioning scheme for the diffusion model to enable both blind and non-blind image restoration. A guidance parameter is also introduced to allow for a trade-off between restoration accuracy and image quality. 
Extensive experiments show that our method significantly outperforms previous approaches on the proposed datasets and real-world artifact images. 
\end{abstract}

\begin{teaserfigure}
  \includegraphics[width=\textwidth]{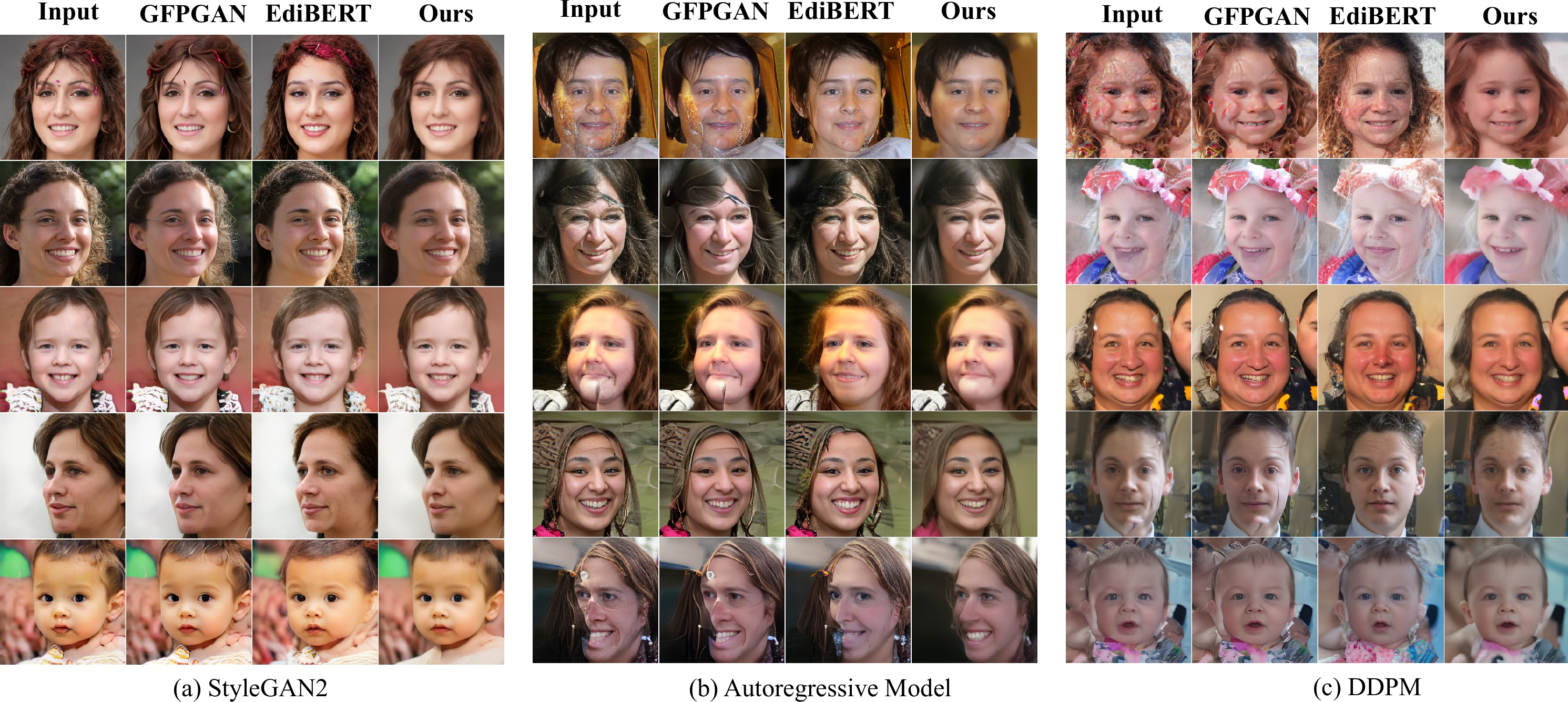}
  \caption{\textbf{Comparisons of GFPGAN~\cite{wang2021towards}, EdiBERT~\cite{Issenhuth2021EdiBERTAG}, and our DiffGAR restoration results on \textit{real image artifacts} generated by different generative models (StyleGAN2~\cite{Karras2019stylegan2}, Autoregressive Model~\cite{esser2021taming}, DDPM~\cite{choi2022perception})}. DiffGAR (the last column) trained with synthetic artifacts dataset can successfully remove the real-world artifacts produced by various generative models.} 
  \label{fig:teaser}
\end{teaserfigure}

\maketitle
\pagestyle{plain} % removes running headers

\section{Introduction}

Generative models such as generative adversarial networks~\cite{Karras2019stylegan2}, autoregressive models~\cite{ostrovski2018autoregressive}, diffusion models~\cite{ho2020denoising}, variational autoencoders~\cite{van2017neural}, and normalization flows~\cite{kingma2018glow}, show impressive capability in photo-realistic image synthesis and editing in recent years. However, due to the imperfect learning of the mapping between distributions or the discontinuity in the latent space, artifacts usually inevitably appear in some of the generation results, as shown in the first column of each subgraph in Fig.~\ref{fig:teaser}. These artifacts may reduce the generation performance and the performance of downstream tasks.

The generative artifacts can be viewed as a special form of image degradation with respect to the implicit \textit{original} image. However, different from traditional image degradation modes~\cite{kawar2022denoising},~\cite{wang2021towards},~\cite{wang2021real} such as JPEG artifacts, Gaussian blur, Gaussian noise and downsampling artifacts, which can usually be represented as a low-order transformation function (e.g., $\bf{y} = [(\bf{x} \otimes  \bf{k}) \downarrow _s + \bf{n}]_{JPEG} $), generative artifacts are non-linear, model-specific, and the transformation function is highly complex, which cast challenges in the modeling of the restoration process. As far as we know, only a few works have been proposed to deal with the artifacts of generative models. 
Two works ~\cite{jeong2022unsupervised},~\cite{tousi2021automatic} have been conducted to detect artifacts produced by GAN models. However, these models are model-specific since they are limited to the latent space of GAN. EdiBERT~\cite{Issenhuth2021EdiBERTAG} learns a masked token prediction model on discredited images, and it assumes that the artifact regions correspond to lower-probability tokens. It achieves image restoration by repredicting those tokens with low probabilities. Although these algorithms can restore images from a specific type of generative artifacts, they are typically model-specific and cannot generalize to different types of generative models.

In this paper, we would like to develop a model-agnostic general framework for the faithful restoration of clean images from various types of generative artifacts. 
It is worth noting that the causes of various types of artifacts generated by different generative models vary by model structure, and it is difficult to propose a general method to directly avoid the artifacts in the generation process.
Therefore, the focus of this work is not to analyze the reasons why the model produces artifacts during generation, but to propose a model-agnostic artifact removal method to improve the quality of image generation.
The term \textit{model-agnostic} denotes that we have no access to the detailed implementation of the generator, but rather consider it as a black-box module. To achieve this, we consider generative artifact removal as an image-to-image translation task, where the only input is the degraded image with generative artifacts, while the output is the restored image. 

One challenge is that no \textit{groundtruth} is available for each degraded image. To generate training data for our restoration model, we design a set of mechanisms to degrade real images that simulate different types of generative artifacts.
Specifically, by exploring the components or parameters that may lead to generative artifacts, we synthesize four generative artifacts from clean images, corresponding to three popular generative models, GAN, autoregressive model and diffusion model. 
Notably, it is inevitable that there is a certain gap between the artifact we build and the real generative artifact produced by generative models.
However, we will demonstrate in the experimental part that the model trained on the synthetic artifact dataset also shows great restoration ability to the real-world generative artifacts.

% baseline model
Next, we propose a baseline method based on a conditional diffusion model, called \textbf{DiffGAR}, for this generative artifact removal task. Given an artifact image and a source image, our algorithm can generate a clean image removing artifacts. 
Traditional degradation restoration tasks aim to estimate the degradation kernel and then produce a clean image through the inversion process~\cite{shocher2018zero}.
Inspired by the recently proposed conditional image-to-image diffusion model~\cite{saharia2021palette}, we treat this artifact removal task as a task to learn the conditional distribution of original clean images given input artifacts, using diffusion models can capture the multi-modal distributions in the high-dimensional image spaces.
We then train our algorithm on our newly proposed artifact-image pairs dataset. 
% blind artifacts removal
Besides, among the traditional degradation restoration tasks, many works focus on solving blind image restoration tasks~\cite{el2020blind}.
Blind artifacts removal aims to restore the original images suffering from unknown degradation kernels or noise levels. 
Similarly, we design a conditioning scheme for the diffusion model to enable blind artifact removal with an extra blind class token embedding.

The overall contributions of this work can be summarized as follows:
\begin{itemize}
    \item We design methods for synthesizing four types of generative artifacts that are as close as possible to real artifacts for restoration model training.
    \item We propose a baseline algorithm, DiffGAR, based on the conditional diffusion model, which achieves high quality restoration from artifact images and shows better reconstruction accuracy and image fidelity than previous works.
    \item We conduct thorough experiments on different settings of the modeling of DiffGAR, including both blind and non-blind restoration mechanisms. We also support a flexible trade-off between image quality and consistency of the results.
    \item The proposed DiffGAR model trained on the synthetic artifact datasets can restore real image artifacts and achieve higher human visual preference than previous works, making it more practical for real-world applications.  
\end{itemize}    

\section{Related Work} \label{Related Work}

\subsection{Image Generation} 

Deep generative models have made great progress in image synthesis tasks recently~\cite{Dhariwal2021DiffusionMB},~\cite{goodfellow2014generative},~\cite{Kingma2014AutoEncodingVB},~\cite{Parmar2018ImageT},~\cite{Song2019GenerativeMB},~\cite{Vahdat2020NVAEAD}.
GAN-based methods have shown amazing capabilities in generating high-fidelity samples, but they have poor mode coverage.
Autoregressive models cover data manifold faithfully, but they often suffer from low sample quality.
Concurrently, the diffusion models have emerged as a promising generative modeling framework, advancing the development of image, audio, and video generation tasks.
However, artifacts are often generated due to architectural limitations of the generative model itself and its inability to capture the complete data manifold pattern~\cite{zhang2019detecting},~\cite{jeong2022unsupervised},~\cite{tan2021systematic}, highlighting the importance of generative artifacts removal for producing more visually appealing images.

\subsection{Traditional Degradation Restoration} 
Many traditional tasks in image restoration can be transformed into linear inverse problems, such as super-resolution, colorization, deblurring, and compressive sensing.
GFPGAN~\cite{wang2021towards} incorporate GAN prior to traditional blind facial image restoration.
~\cite{wang2021real} trains a practical Real-ESRGAN for real-world blind super-resolution with pure synthetic training pairs.
Besides, diffusion models have recently been introduced into the image restoration tasks.
Denoising Diffusion Restoration Models (DDRM)~\cite{kawar2022denoising} has been proposed as a general linear inverse problem solver based on unconditional diffusion generative models.

\subsection{Generative Artifacts Restoration} 
Recently, there are a few works focusing on exploring the unique artifacts in GAN model architectures.
The author ~\cite{jeong2022unsupervised} removes artifacts through ablating units that are related to artifact generations.
The paper ~\cite{tan2021systematic} propose a novel pixel-instance normalization (PIN) layer to remove the circular artifacts for vanilla StyleGAN.
However, these methods are specifically designed for partial GAN artifacts.
EdiBERT~\cite{Issenhuth2021EdiBERTAG} is a kind of generative model which models the VQGAN latent space in a non-autoregressive manner. EdiBERT replaces the autoregressive GPT model in Taming Transformers with a bi-directional BERT~\cite{devlin2018bert} model.
When applied to image denoising tasks,  EdiBERT will first detect discrete tokens in a given artifact image that do not fit appropriately in the sequence $s$, and then change these tokens to new tokens so that the likelihood of the new sequence can be higher.

\section{Synthesis of Generative Artifacts} \label{Dataset}

In this section, we describe our method of synthesizing different kinds of generative artifacts corresponding to three typical types of generative models, GAN, Autoregressive model and DDIM. Furthermore, we synthesize training pairs using images from FFHQ dataset~\cite{karras2019style} and AFHQ-Dog dataset~\cite{choi2020stargan}.

\subsection{GAN Artifacts}

Since it is difficult to directly synthesize high-quality, id-preserving artifacts-image pairs based on StyleGAN2~\cite{Karras2019stylegan2}, we use VQGAN~\cite{esser2021taming} instead, which is also a GAN model.
Taming Transformers~\cite{esser2021taming} is a two-stage generative model that first learns a latent representation of the data and then, in a second stage, an autoregressive gpt probabilistic model of the latent representation.
VQGAN is a vector-quantized image model which can map an image into a sequence of discrete latent variables, serving as the first stage model of Taming Transformers. 
The vector-quantized image model consists of an encoder $E$, a decoder $D$ and a codebook $\mathcal{Z} = \left \{ \bm{z}_k \right \} _{k=1}^{K} \in \mathbb{R}^{k \times d} $, where $K$ is the number of discrete codes in the codebook $Z$ and $d$ is the dimension of codes. 
The convolutional encdoer $E$ downsamples an image $ \bm{x} \in \mathbb{R}^{c\times H \times W}$ into a feature map $ \bm{z}_e \in \mathbb{R}^{d\times h \times w}$.
Then each spatial feature vector $ \bm{z}_{ij}$ is substituted via a nearest-neighbour lookup onto a discrete codebook entry $ \bm{z}_k$:

\begin{equation}
  \bm{z}_q = \left( argmin_{\bm{z}_k \in \mathcal{Z}} \|\bm{z}_{ij} - \bm{z}_k\|_2^2 \right) \in \mathbb{R}^{d\times h \times w} ~.
    \label{equ:quantize}
\end{equation}
where $h \times w$ represents the sequence length of the quantized image. 
Subsequently, the quantized encoding $\bm{z}_q$ is fed to the decoder $D$ to reconstruct the input $\tilde{x} = D(\bm{z}_q)$.

For VQGAN artifacts, given an image $x$, we can obtain the image tokens sequence $s = (s_1, \cdots , s_{h \times w} \in \mathbb{Z}^{1 \times h \times w} )$ through a well-trained VQGAN model.
Then, we randomly select a rectangle mask within which the image tokens will be replaced by a randomly sampled tokens sequence. 
In this case, most of the image information can be preserved, while obvious artifacts will be caused in the local area of the image.

\subsection{Autoregressive Model Sampling Artifacts}

The second stage model of Taming Transformers~\cite{esser2021taming} is a transformer-based autoregressive generative model. 

With encoder $E$ and decoder $D$ available, we can obtain discrete tokens $s = (s_1, s_2, \cdots, s_l)$ for a given image. 
In the second stage of Taming Transformers, for a given discrete token sequence $s = (s_1, s_2, \cdots, s_l)$, an autoregressive transformer is trained to predict the next sequence token conditioned on the previously predicted image tokens:
\begin{equation}
    p_{\theta}(s) = \prod_{i=1}^{l} p_{\theta}^i(s_i|s_{<i}) ~.
    \label{equ:GPT}
  \end{equation}

During inference, the most commonly used sampling strategies of autoregressive models are temperature sampling and top-k sampling.
We can implement temperature sampling by dividing the logits outputs of the autoregressive transformer model by the temperature, which is then fed into a softmax layer and outputs the sampling probabilities of tokens in the vocabulary.
Higher temperatures make the model decreasingly confident in its top choices. 
The top-k sampling means ordering by probability and zeroing out the probabilities below the kth token.
We simulate autoregressive model sampling artifacts via a higher temperature and top-k value.
In the remaining part of the paper, we use GPT sampling artifacts to represent the autoregressive model sampling artifacts for simplicity.

\begin{figure*}
  \begin{center}
  \includegraphics[width=0.9 \linewidth]{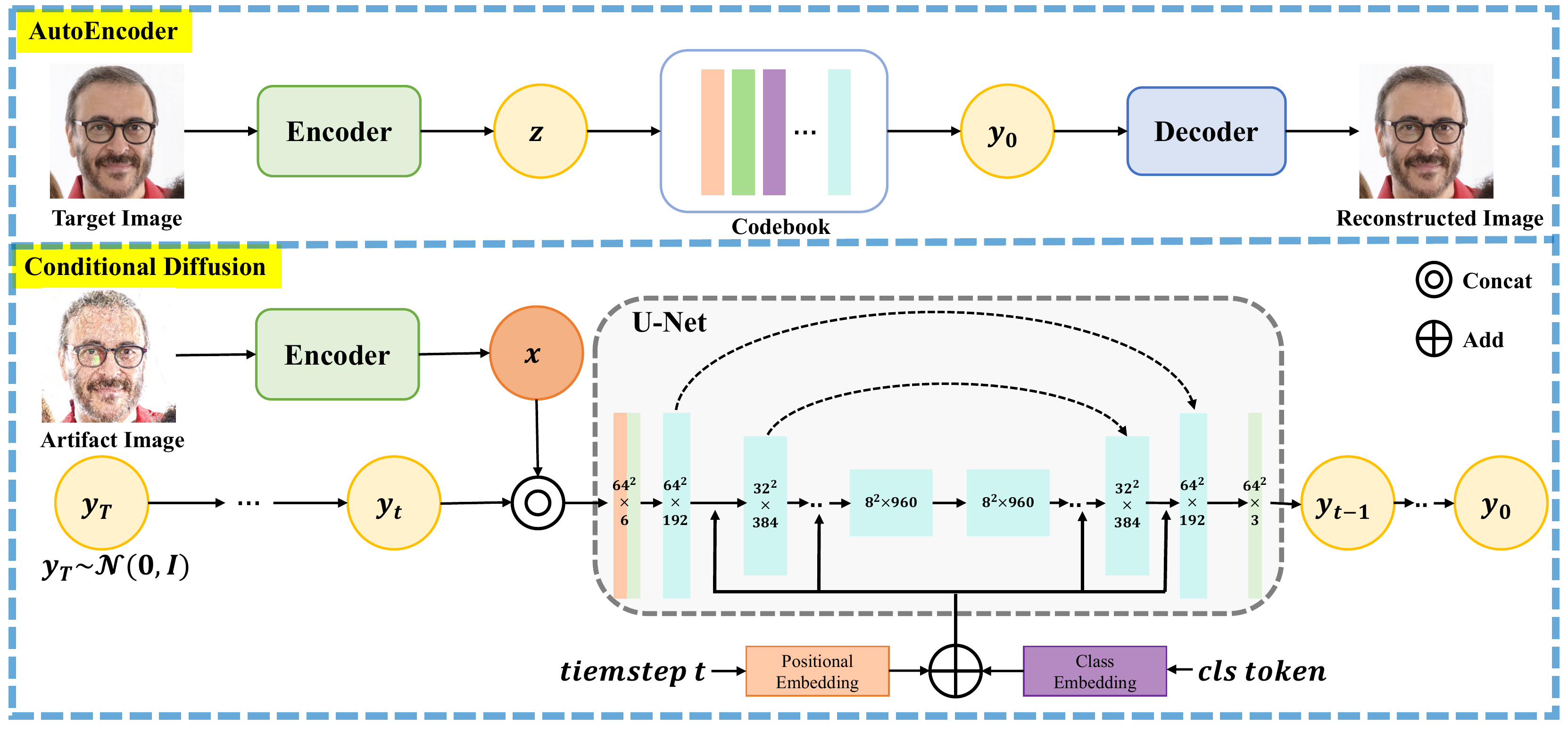}
  \end{center}
  \caption{\textbf{Network architecture of DiffGAR.} DiffGAR includes two components: an autoencoder(top) and a conditional diffusion model(bottom). The autoencoder maps the input image into a latent representation. During training, the latent representation of  artifact image and target image are concat as input to the U-Net network.}
  \label{fig:model_architecture}
\end{figure*}

\subsection{DDIM Artifacts}  \label{sec:ddim_artifact}

Recently, the Denoising Diffusion Probabilistic Model(DDPM) has achieved great success in image generation, audio synthesis~\cite{chen2020wavegrad},~\cite{kong2020diffwave}, image super-resolution~\cite{kadkhodaie2020solving},~\cite{saharia2021image} and image editing~\cite{meng2021sdedit},~\cite{sinha2021d2c} tasks.
DDPM is a latent variable generative model which consists of a forward diffusion process and a reverse diffusion process.
The forward diffusion process gradually adds noise to the data via a fixed Markov chain through $T$ steps and yields a sequence of increasingly noisy images $\bm{x}_1, \cdots , \bm{x}_T$ which share the same dimensionality as a given image $\bm{x}_0$. 
Each step in the forward diffusion process is assumed to fit a Gaussian distribution $q(\bm{x}_t|\bm{x}_{t-1}) := \mathcal N(\sqrt{1-\beta_t}\bm{x}_{t-1}, \beta_t \mathbf{I})$, where $\beta_t$ is the hyperparameter of the noise schedule.
Note that, due to the property of the Markov chain, we can marginalize the forward process and derive the latent variable $\bm{x}_t$ at arbitrary timestep directly from $\bm{x}_0$ as:
\begin{equation}
    \bm{x}_t = \sqrt{\bar{\alpha}_t } \bm{x}_0 + \sqrt{1-\bar{\alpha}_t }  \bm{\epsilon}, \bm{\epsilon} \sim  \mathcal N(0, \mathbf{I}) ~.
    \label{equ:diffusion_forward}
  \end{equation}

where $\bar{\alpha}_t := \prod_{i=1}^{t} (1 - \beta_i)$. The reverse process is also assumed to be a Markov chain with a Gaussian distribution $p_{\theta}(\bm{x}_{t-1}|\bm{x}_t) := \mathcal N(\bm{x}_{t-1};\mu_{\theta}(\bm{x}_t,t), \Sigma_{\theta}(\bm{x}_t,t))$ that gradually predicts a denoised variant of the input sample to the desired output.
$\mu_{\theta}(\bm{x}_t,t)$ can be decomposed into a linear combination of $\bm{x}_t$ and a noise diffusion model $\epsilon_{\theta}(\bm{x}_t,t)$:
\begin{equation}
  \mu_{\theta}(\bm{x}_t, t) = \frac{1}{\sqrt{\alpha_t}} (\bm{x}_t - \frac{\beta_t}{\sqrt{1-\bar{\alpha}_t}}\epsilon_{\theta}(\bm{x}_t,t)) ~.
  \label{equ:mu_epsilon}
\end{equation}
The noise diffusion model is trained by minimizing the variational upper bound on the negative log-likelihood over the training data, and the following is a simplified training objective from ~\cite{ho2020denoising}:
\begin{equation}
  \mathcal{L}_{simple} = \mathbb{E}_{t,\bm{x}_0, \bm{\epsilon}} \left \| \bm{\epsilon} - \bm{\epsilon}_{\theta}(\bm{x}_t, t)  \right \|_2^2  ~.
  \label{equ:simple_loss_func}
\end{equation}

DDIM~\cite{song2020denoising} parameterizes the forward process as a non-Markovian and deterministically maps noise back to the original data samples during inference:
\begin{equation}
  \bm{x}_{t-1} =  \sqrt{\bar{\alpha}_{t-1}} \bm{x}_0 + \sqrt{1-\bar{\alpha}_{t-1} - \sigma_t^2} \bm{\epsilon}_t + \sigma_t \bm{\epsilon} ~.
  \label{equ:ddim_sample}
\end{equation}
by setting the noise variance $\sigma_t$ to $0$, we can obtain the deterministic DDIM sampling process, enabling the inversion of a given latent code into its original image.

Following~\cite{kim2021diffusionclip}, the deterministic forward DDIM process to obtain a latent code is represented as:
\begin{equation}
    \bm{x}_{t+1} = \sqrt{\bar{\alpha}_{t+1}}\tilde{\bm{x}}_0 + \sqrt{1-\bar{\alpha}_{t+1}}\epsilon _{\theta } (\bm{x}_t,t) ~.
    \label{equ:ddim_forward}
  \end{equation}
Given an image $\bm{x}$ and the stopping step $T_0$ of the forward process, we can obtain the unique latent code $\bm{x}_{T_0}$ of that image.
Besides, the deterministic inverse DDIM sampling process from the obtained latent code is as follows:
\begin{equation}
    \bm{x}_{t-1} = \sqrt{\bar{\alpha}_{t-1}}\tilde{\bm{x}}_0 + \sqrt{1-\bar{\alpha}_{t-1}}\epsilon _{\theta } (\bm{x}_t,t) ~.
    \label{equ:ddim_reverse}
  \end{equation}
where $\tilde{\bm{x}}_0 = \frac{\bm{x}_t - \sqrt{1-\bar{\alpha}_t}\epsilon_{\theta}(\bm{x}_t,t)}{\sqrt{\bar{\alpha}_t}}$.
We construct two types of artifacts related to DDIM sampling process. 
Since the initial latent code is the basis for the whole subsequent generation process, it has a direct impact on the image generation quality. 
Therefore, we construct artifact-image pairs by perturbing the initial latent code.
Firstly, we perturb the obtained latent code with additive white Gaussian noise, $\tilde{\bm{x}}_{T_0} = \bm{x}_{T_0} + \alpha * \bm{n}, \bm{n} \sim \mathcal N(0, \mathbf{I})$, where $\alpha$ is a hyperparameter controlling the strength of the added noise.
Secondly, we perform scaling and shifting operation using the latent code $\tilde{\bm{x}}_{T_0} = \gamma * \tilde{\bm{x}}_{T_0} + \beta$, where $\gamma$ and $\beta$ correspond to the scale and shift coefficients.
These two operations sometimes result in streaky artifacts in the image.

\section{Conditional Diffusion Models}
\label{sec:method}

We utilize the diffusion model as the generative artifact removal backbone due to its powerful generative capability.
Following Latent Diffusion~\cite{rombach2022high}, to reduce the computational cost and fitting difficulty of training a diffusion model, we use a two-stage diffusion model as our method.
In the first stage, an autoencoder model compresses the input image into a low-dimensional latent space.  We then construct a powerful conditional diffusion model over the compressed latent space to learn the joint distribution.
Fig.~\ref{fig:model_architecture} shows the structure of our model DiffGAR.

\noindent\paragraph{Stage One-AutoEncoder Model.} Based on~\cite{esser2021taming} which learns a discrete latent space, we learn a continuous compressed latent space which allows training a continuous diffusion model.
Given an image $\bm{x} \in \mathbb{R}^{3\times H \times W}$ in RGB space, the encoder $E$ first encodes the image into a latent representation $\bm{z} \in \mathbb{R}^{d\times h \times w}$, then $\bm{z}$ can be fed into the decoder $D$ to reconstruct the input image.

\noindent\paragraph{Stage Two-Conditional Diffusion Model.} Given the generative artifacts dataset of artifact-image pairs, denoted $D = \{X_i, Y_i \}_{i=1}^{N}$, the goal of our method is to learn a parametric approximation to $p(Y|X)$ that maps a source artifact image $X$ to the clean target image $Y$.
With the trained first-stage autoencoder, we can obtain the latent representation  $\bm{x}$, $\bm{y}_0$ of a given artifact image and a clean target image. 
Compared to the original high-dimensional image space, the compressed low-dimensional pixel space is more efficient to train the conditional diffusion model.
Taking the source artifact image $\bm{x}$ and the noisy target image $\bm{y}_t$ as input, we can train a neural backbone $\epsilon_{\theta}$ to progressively denoise $\bm{y}_t$ and produce the target image $\bm{y}_0$. 
During training, we use a shared U-Net~\cite{ronneberger2015u} architecture for all timesteps $t$, by injecting the current timestep $t$ with sinusoidal position embedding~\cite{vaswani2017attention}.
Additionally, since the artifact images in the proposed dataset have different artifact class labels, we can utilize the class labels as additional information to guide the diffusion process.
Specifically, we learn a class embedding matrix for the proposed four types of generative artifacts during training and add the class embedding to the timestep embedding for each sample during training.
However, similar to the blind image restoration tasks where the degradation model is not provided, when applied to artifact images generated directly from the generative models, we cannot know exactly which of the four artifacts in the dataset the current artifact image belongs to.
Therefore, we learn an extra $[\text{\tt{MASK}}]$ class token $cls_m$ which represents the unknown artifact class then it can be applied to blind image restoration.
During training, we randomly set $50\%$ of the input artifact class tokens to the $[\text{\tt{MASK}}]$ class token. The detailed training algorithm is shown in  Algorithm~\ref{alg:train_model}.

\begin{algorithm}
  \caption{Training a denoising model $\epsilon_{\theta}$.}\label{alg:train_model}
  \begin{algorithmic}[1] 
  \Repeat
  \State $(X, Y, cls) \gets $ sample training artifact-image pair and artifacts class token
  \State $\bm{x} \gets E(X), \bm{y}_0 \gets E(Y)$
  \State $t \gets $ Uniform$ (\{ 1, \cdots , T\})$
  \State $\bm{\epsilon} \sim  \mathcal N(\mathbf{0}, \mathbf{I}) $
  \State $ \mathcal{L} \gets \left \| \epsilon_{\theta }(\bm{x}, \sqrt{\bar{\alpha}_t }\bm{y}_0+\sqrt{1-\bar{\alpha}_t }\bm{\epsilon}, t | cls) - \bm{\epsilon}   \right \|_2^2 $
  \State $ \theta \gets \theta - \eta \Delta_{\theta}\mathcal{L} $
  \Until{converged}
  \end{algorithmic}
\end{algorithm}

\noindent\paragraph{Blind / Non-Blind Generative Artifacts Restoration.} In the inference stage, we accelerate the generative process by using fewer discretization steps~\cite{song2020denoising}.
Besides, we design three kinds of generative artifacts restoration methods, non-blind restoration, blind restoration and a tradeoff between them.
For non-blind image restoration, we set the artifact class token to the true artifact class.
For blind image restoration, the $[\text{\tt{MASK}}]$ class token is fed into the noise diffusion model. 
Besides, inspired by the classifier-free guidance sampling in previous conditional image generation tasks~\cite{ho2021classifier},~\cite{nichol2021glide} which can improve the generation quality, we adapt this strategy to our generative artifacts restoration task by introducing classifier-free guidance sampling conditioned on artifact class token.
During sampling, the output of the diffusion model is further extrapolated in the direction of true class token input $\epsilon_{\theta}(\bm{x}, \bm{y}_t, t|cls_r)$ and away from the $[\text{\tt{MASK}}]$ class token input $\epsilon_{\theta}(\bm{x}, \bm{y}_t, t|cls_m)$:
\begin{equation}
  \begin{split}
  \hat{\epsilon }_{\theta}(\bm{x}, \bm{y}_t, t|cls_r) = \epsilon_{\theta}(\bm{x},\bm{y}_t, t|cls_m) \\  +  s(\epsilon_{\theta}(\bm{x}, \bm{y}_t, t|cls_r) - \epsilon_{\theta}(\bm{x}, \bm{y}_t, t|cls_m))  ~.
  \label{equ:classifier_guidance}
  \end{split}
\end{equation}
Here $s \ge 1$ is the guidance scale. Algorithm~\ref{alg:inference_model} provides the detailed sampling process.
Given an artifact image, the model first samples the noise from a Gaussian distribution, then the model takes the artifact image, the artifact class token, and the image produced from the previous time step as input, and gradually denoises it to generate a clean target image.

\begin{algorithm}
  \caption{Inference of the denoising model, given fast inference time stride $\triangle_t$, input artifact image $X$.}\label{alg:inference_model}
  \begin{algorithmic}[1] 
  \State $t \gets T, \bm{x} \gets E(X), cls_r$ is the artifacts class token of $X$ 
  \State $ \bm{y}_t \gets \mathcal N(\mathbf{0}, \mathbf{I}) $
  \While{$t > 0$}
  \State $ \bm{\epsilon} \gets \hat{\epsilon}_{\theta}(\bm{x}, \bm{y}_t, t | cls_r)$
  \State $ \tilde{\bm{y}}_0 = \frac{\bm{y}_t - \sqrt{1-\bar{\alpha}_t}\bm{\epsilon}}{\sqrt{\bar{\alpha}_t}}$
  \State $\bm{y}_t = \sqrt{\bar{\alpha}_{t-1}} \tilde{\bm{y}}_0 + \sqrt{1-\bar{\alpha}_{t-1}} \bm{\epsilon}$
  \State $t \gets t - \triangle_t$
  \EndWhile
  \end{algorithmic}
  \Return $D(\bm{y}_t)$
\end{algorithm}

\begin{table*}[]
  \caption{\textbf{Quantitative restoration results on synthetic datasets between GFPGAN (\cite{wang2021towards}), Real-ESRGAN (\cite{wang2021real}), EdiBERT (\cite{Issenhuth2021EdiBERTAG}) and Ours.} The comparison results show that our method produces higher quality output images.}
 \renewcommand\arraystretch{1.2} 
  % \footnotesize
  {\scalebox{0.7}{
  \begin{tabular}{c|c|cc|cc|cc|cc|cc}
  \hline
  \multirow{2}{*}{Metric}         & \multirow{2}{*}{Method} & \multicolumn{2}{c|}{Replace Token} & \multicolumn{2}{c|}{GPT Sampling} & \multicolumn{2}{c|}{DDIM Gaussian} & \multicolumn{2}{c|}{DDIM Scale} & \multicolumn{2}{c}{ALL} \\ \cline{3-12} 
  &                         & FFHQ            & AFHQ             & FFHQ             & AFHQ             & FFHQ             & AFHQ            & FFHQ            & AFHQ            & FFHQ             & AFHQ               \\ \hline
  \multirow{4}{*}{FID$\downarrow$}  & GFPGAN                   & 19.32   &                         & 22.1   &                         & 28.41   &                         & 45.93  &                         & 25.5                               &                               \\
                           & Real-ESRGAN              &         & 27.43                   &        & 28.98                   &         & 33.95                   &        & 36.51                   &                          & 29.5                          \\
                           & EdiBERT                  & 16.56   & 33.52                   & 16.95  & 42.81                   & 34.13   & 44.96                   & 43.46  & 45.96                   & 22.74                         & 43.34                         \\
                          & \textbf{DiffGAR(Ours)}                   & \textbf{15.15}   & \textbf{27.22}                   & \textbf{16.72}  & \textbf{28.97}                   & \textbf{20.17}   & \textbf{27.45}                   & \textbf{25.05}  & \textbf{29.99}                   & {\color[HTML]{FF0000} 18.34}  & {\color[HTML]{FF0000} 28.48}  \\ \hline
 \multirow{4}{*}{MSE$\downarrow$}                  & GFPGAN                   & 0.0384  & {\color[HTML]{FF0000} } & 0.0472 & {\color[HTML]{FF0000} } & 0.0131  & {\color[HTML]{FF0000} } & 0.0577 & {\color[HTML]{FF0000} } &        0.041                        &                               \\
                           & Real-ESRGAN              &         & 0.0412                  &        & 0.0494                  &         & 0.0228                  &        & 0.038                   &                         & 0.0367                        \\
                           & EdiBERT                  & 0.0701  & 0.066                   & 0.0734 & 0.0781                  & 0.0763  & 0.0715                  & 0.1329 & 0.0715                  & 0.081                         & 0.0692                        \\
     & \textbf{DiffGAR(Ours)}                   & \textbf{0.0336}  & \textbf{0.0355}                  & \textbf{0.0456} & \textbf{0.044}                   & \textbf{0.0122}  & \textbf{0.0196}                  & \textbf{0.0531} & \textbf{0.0301}                  & {\color[HTML]{FF0000} 0.036}  & {\color[HTML]{FF0000} 0.0331} \\ \hline
     \multirow{4}{*}{PSNR$\uparrow$}                   & GFPGAN                   & 20.585  & {\color[HTML]{FF0000} } & 19.734 & {\color[HTML]{FF0000} } & 25.451  & {\color[HTML]{FF0000} } & 18.585 & {\color[HTML]{FF0000} } &      21.02                           &                               \\
                           & Real-ESRGAN              &         & 20.17                   &        & 19.36                   &         & 22.81                   &        & 20.9                    &                        & 20.93                         \\
                           & EdiBERT                  & 18.053  & 18.02                   & 17.868 & 17.34                   & 17.683  & 19.67                   & 15.101 & 18.67                   & 17.17                         & 17.94                         \\
     & \textbf{DiffGAR(Ours)}                  & \textbf{21.393}  & \textbf{20.89}                   & \textbf{20.084} & \textbf{19.965 }                 &\textbf{ 25.744}  & \textbf{23.49}                   &\textbf{ 18.921} & \textbf{21.71}                   & {\color[HTML]{FF0000} 21.48}  & {\color[HTML]{FF0000} 21.45}  \\ \hline
    \multirow{4}{*}{SSIM$\uparrow$}                 & GFPGAN                   & 0.6101  &                         & 0.584  &                         & \textbf{0.775}   &                         & 0.677  &                         &     0.658                           &                               \\
                           & Real-ESRGAN              &         & 0.5189                  &        & 0.48                    &         & \textbf{0.662 }                  &        &\textbf{ 0.6609}                  &                         & 0.576                         \\
                           & EdiBERT                  & 0.5415  & 0.4621                  & 0.535  & 0.428                   & 0.442   & 0.52                    & 0.395  & 0.421                   & 0.481                         & 0.458                         \\
    & \textbf{DiffGAR(Ours)}                  & \textbf{0.6182}  & \textbf{0.5571}                  & \textbf{0.593}  &\textbf{ 0.541}                   & 0.767   & 0.6484                  &\textbf{ 0.683}  & 0.614                   & {\color[HTML]{FF0000} 0.665}  & {\color[HTML]{FF0000} 0.587}  \\ \hline
    \multirow{4}{*}{\makecell{ID \\ Consistency} $\uparrow$}                  & GFPGAN                   & \textbf{0.6169}  &                         &\textbf{ 0.5852 }&                         & \textbf{0.8727}  &                         & \textbf{0.771}  &                         &      {\color[HTML]{FF0000} 0.7143}                          &                               \\
                           & Real-ESRGAN              &         & 0.9019                  &        & 0.8966                  &         & 0.8982                  &        & 0.8996                  & & 0.8984                        \\
                           & EdiBERT                  & 0.5027  & 0.9312                  & 0.4989 & 0.9192                  & 0.5231  & 0.913                   & 0.4588 & 0.912                   & 0.4954                        & 0.9165                        \\
      & \textbf{DiffGAR(Ours)}                  & 0.5565  & \textbf{0.9496}                  & 0.5007 & \textbf{0.9455 }                 & 0.8411  & \textbf{0.9513 }                 & 0.7268 &\textbf{ 0.9421  }                & 0.6536                        & {\color[HTML]{FF0000} 0.9467} \\ \hline
  \end{tabular}
  }}
  \label{tab:compare_quantiative}
\end{table*}

\section{Experiments}
\label{sec:experiments}

\begin{figure*}
  \begin{center}
  \includegraphics[width=0.9 \linewidth]{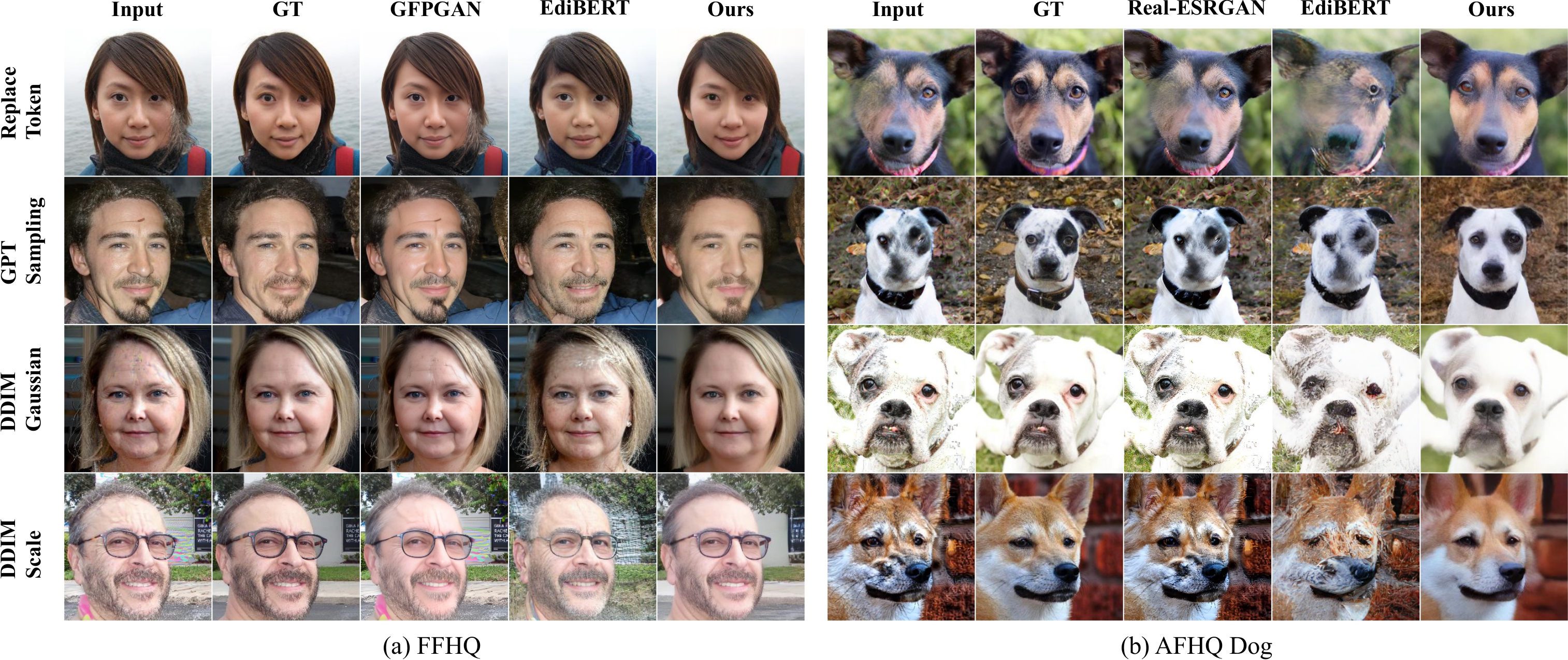}
  \end{center}
  \caption{\textbf{Restoration results comparison on proposed FFHQ artifacts dataset (a) and AFHQ-Dog artifacts dataset (b).} Each row shows the different types of generative artifacts.}
  \label{fig:compare_vis}
\end{figure*}

We evaluate the effectiveness of our DiffGAR model on the proposed synthetic generative artifact images as well as the real artifact images sampled from the generative models.
We compare DiffGAR with two traditional restoration models GFPGAN~\cite{wang2021towards} (for FFHQ artifact dataset), Real-ESRGAN~\cite{wang2021real} (for AFHQ-Dog artifact dataset) and a generative artifacts removal model EdiBERT~\cite{Issenhuth2021EdiBERTAG}.

\noindent\paragraph{Datasets Construction.} We generate artifact images on the FFHQ and AFHQ-Dog dataset at $256 \times 256$ resolution. 
For the FFHQ dataset, we use 68k images for training and 2k images for inference. Here are some parameter settings for the training dataset. For the AFHQ-Dog dataset, 4735 images are used for training, and 500 images are used for inference.
For GPT sampling artifacts, we use the well-trained model from~\cite{esser2021taming} and set different temperatures and top-k values to simulate GPT sampling artifacts.
For the FFHQ dataset, we set the number of top-k to 500 and the sampling temperature to 21. 
For the AFHQ-Dog dataset, we set the number of top-k to 600 and the sampling temperature to 22. 
Notably, in order to keep the identity of the object in the image, we set $90\%$ of image tokens unchanged and only $10\%$ of the tokens will be resampled using the given temperature and top-k. 
For replace\_token artifacts, we randomly sample a $4 \times 4$ rectangle and replace the tokens within the selected rectangle with randomly sampled tokens from the codebook.
% When generating ddim\_recons artifacts, we set the stopping step of forward process $T_0$ to $500$, forward diffusion steps to $40$, inversion steps to $8$ for both datasets.
When generating ddim\_gaussian artifacts, we set the stopping step of the forwarding process $T_0$ to $840$, forward diffusion steps and inversion steps to $48$, $\alpha$ to 0.3. 
Ddim\_scale artifacts share the same diffusion steps as ddim\_gaussian, and $\gamma$, $\beta$ is set to be $1.015$, $0.01$ for the FFHQ dataset.
For AFHQ-Dog, we set $\gamma$ to $1.015$, $\beta$ to $0.001$.

\noindent\paragraph{Evaluation Metrics.}  We use FID~\cite{heusel2017gans} as the generation quality comparison metric.
The image reconstruction quality is evaluated by mean squared error (MSE) in image pixel level space.
Besides, we adopt pixel-wise metrics PSNR and SSIM.
To measure the identity preserving ability, we use iResNet~\cite{behrmann2019invertible} and CLIP~\cite{radford2021learning} to extract image features for images in FFHQ and AFHQ-Dog datasets, respectively, and then calculate the similarity between features.

\noindent\paragraph{Training Details.} We use the pretrained model of an autoencoder network from Latent-diffusion~\cite{rombach2022high}, which maps the input image into a latent representation $\bm{z} \in \mathbb{R}^{3 \times 64 \times 64}$.
We train our DiffGAR model for 390k and 87k training steps on the FFHQ dataset and AFHQ-Dataset dataset with a batch size of 24, respectively.
We adopt a U-Net architecture for the diffusion model and consider $T = 1000$ for the diffusion process. During inference, we use $30$ DDIM sampling timesteps.

\begin{figure*}[!h]
  \centering
  \begin{subfigure}{0.9\linewidth}
    \includegraphics[width=\linewidth%height=9cm, width=9cm
    ]{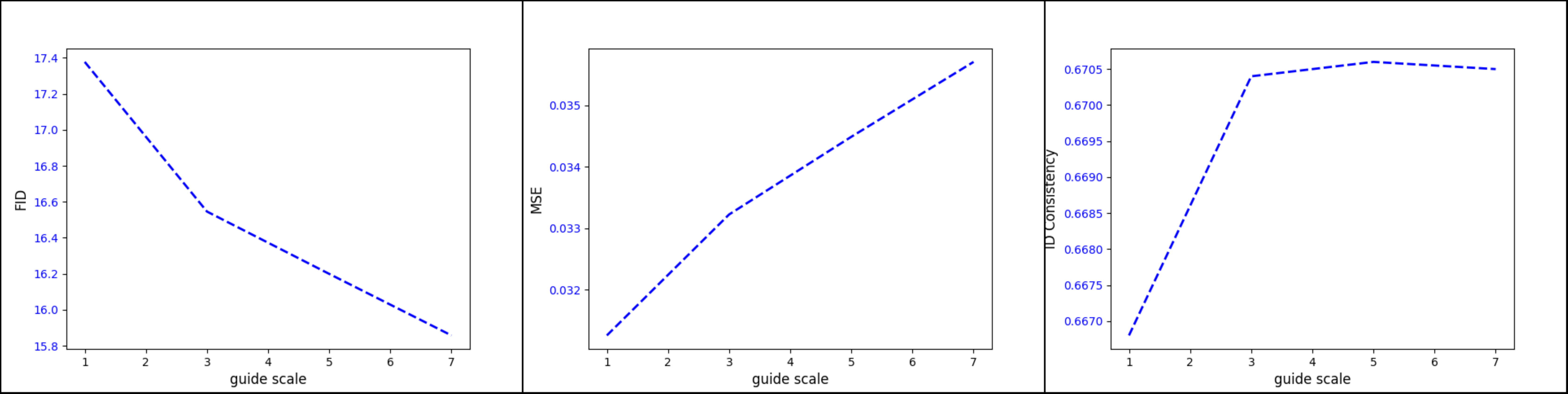} 
    \caption{Classifier-free guidance scale.}
    \label{fig:ablation-guide}
  \end{subfigure}
  \hfill
  \begin{subfigure}{0.9 \linewidth}
    \includegraphics[width=\linewidth %height=9cm, width=9cm
    ]{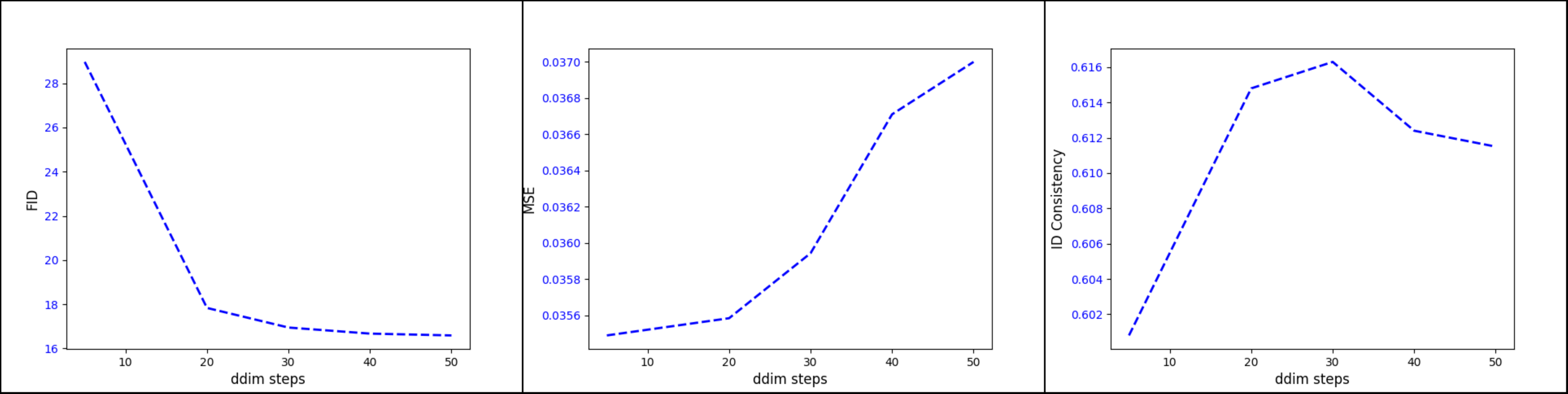}
    \caption{DDIM sampling steps.}
    \label{fig:ablation_ddim_step}
  \end{subfigure}
  \caption{\textbf{Ablation study on the classifier-free guidance scale and the number DDIM sampling steps on the FFHQ dataset.}}
  \label{fig:ablation_all}
\end{figure*}

\subsection{Comparison Results}

\noindent \paragraph{Quantitative Evaluation.} Tab.~\ref{tab:compare_quantiative} shows the quantiative restoration results for the proposed generative artifact datasets.
We used different levels of artifacts in the testing phase compared to the training phase.
We show the overall restoration performance (the last column in Tab.~\ref{tab:compare_quantiative}), that is, for each image, we randomly select one of the four artifacts for restoration.
Our model achieves the best performance on all these five metrics except the ID Consistency metric on FFHQ artifacts dataset.
Although GFPGAN~\cite{wang2021towards} can better preserve the human face identity due to the identity loss applied in the training stage, the restored image quality is far beyond acceptance as the FID score is high ($22.5$).
We also show the performance on each type of synthetic artifact for reference.
Besides, we conduct comparative experiments on real artifacts generated by generative models. 
We generate batches of images from three generative models (StyleGAN2, Autoregressive Model and DDPM) and randomly selected 50 artifact images. For each instance, 10 different evaluators are required to choose between three restored images generated by the three models being compared.
The evaluation criteria involve two aspects: identity preservation and image restoration quality.
The result is provided in Tab.~\ref{tab:real_human}. Compared with previous works, ours achieves high favourability, demonstrating the remarkable capability of restoring real-world generative artifacts. 

\begin{table}
  \caption{\textbf{Comparison with previous work on real-world generative artifacts (human preference).}}
  \begin{center}
    {\scalebox{0.75}{
\begin{tabular}{c|c }
\toprule
\textbf{ Model}   & \textbf{Preference}$\uparrow$   \\
    \midrule
    %  Method& FFHQ& CUB & FFHQ & CUB & FFHQ & CUB &FFHQ& CUB  \\
    GFPGAN  & 27\%  \\
    \midrule   
    EdiBERT & 18\%  \\
    \midrule 
    DiffGAR(Ours) & 55\%  \\
    \bottomrule
\end{tabular}
}}
\end{center}
\label{tab:real_human}
\end{table}

\noindent \paragraph{Qualitative Evaluation.} Qualitative results are presented in Fig.~\ref{fig:compare_vis} and Fig.~\ref{fig:teaser}.
Compared to other methods, DiffGAR produces high quality restoration results, as shown in Fig.~\ref{fig:compare_vis}.
The two traditional image restoration models GFPGAN~\cite{wang2021towards} and Real-ESRGAN~\cite{wang2021real} do not have the ability to restore image regions with semantic errors.
While EdiBERT can remove discrete artifacts such as Replace Token artifacts and GPT Sampling artifacts as shown in the first two rows of Fig.~\ref{fig:compare_vis}, the identity of the person changes slightly,  while our method DiffGAR can preserve the identity better and successfully remove the artifacts.
For the remaining DDIM-related artifacts involving the entire image, only DiffGAR can successfully remove the artifacts.
Fig.~\ref{fig:teaser} shows the restoration results of real artifacts generated by GAN~\cite{Karras2019stylegan2}, GPT~\cite{esser2021taming}, DDPM model~\cite{choi2022perception}, respectively. 
DiffGAR exhibits better image restoration capabilities compared to other models in all three kinds of generative artifacts.

\subsection{Ablation Study}

\begin{table}
  \caption{\textbf{Ablation Study blind/non-blind image restoration.}}
  \begin{center}
    {\scalebox{0.75}{
\begin{tabular}{c|c |c  |c }
\toprule
     & \textbf{FID}$\downarrow$&  \textbf{MSE}$\downarrow$&    \textbf{ID consistency}$\uparrow$ \\
    \midrule
    %  Method& FFHQ& CUB & FFHQ & CUB & FFHQ & CUB &FFHQ& CUB  \\
    \textbf{blind}  & 17.45 & 0.0347    & 0.6295  \\
    \midrule 
    \textbf{non-blind} & 17.79  & 0.0298 & 0.6843 \\
    \midrule 
    \textbf{guidance} & 16.89  & 0.0316  & 0.6886  \\
    \bottomrule
\end{tabular}
}}
\end{center}
\label{tab:ablation_classifier}
\end{table}

We conduct ablation studies on the FFHQ artifact dataset.
\noindent \paragraph{Blind / Non Blind Image Restoration.} 
We investigate how different artifacts restoration methods affect the restoration performance.
As shown in Tab.~\ref{tab:ablation_classifier}, blind image restoration performs the worst in terms of reconstruction quality.
The non-blind image restoration strategy recovers most of the details of the target image with the smallest MSE value.
We find that the classifier-free guidance clearly improves the image quality in terms of FID, at the cost of reduced reconstruction detail.
Fig.~\ref{fig:ablation-guide} shows how the guidance scale $s$ affects the restoration results. 
We find that when $s$ increases, the FID score and ID consistency become better, while the MSE score decreases.
We hypothesize that this may be because the classifier-free guidance pushes the samples toward the true data distribution while ignoring instance details. 
In this paper, we choose a guide scale of 3.0 to achieve a trade-off between image quality and reconstruction accuracy.
However, it is off to the users to choose a proper setting according to their requirements.

\noindent \paragraph{DDIM sampling steps.} We study the impact of the number of DDIM sampling steps on restoration performance. As shown in Fig.~\ref{fig:ablation_ddim_step}, as the number of sampling steps increases, the FID score and ID consistency get better while the MSE score decreases.
When the number of sampling steps increases to 30, the ID consistency score starts to drop.
We assume that the more sampling steps, the better the image quality, but at the cost of changing more details of the image to get closer to the real manifold space.
We use $30$ DDIM sampling steps in our experiment because it provides a good trade-off between various evaluation metrics.

\section{Conclusion}

This paper introduces the generative artifacts restoration task along with two new generative artifact datasets.
Furthermore, we design a general generative artifacts restoration framework DiffGAR based on a conditional diffusion model and demonstrate the great potential of restoring clean images from artifact images.
We expect this work could attract more researchers from the community to further study the artifacts removal of generated images, including a better method of simulating generative artifacts and the development of new methods to improve the image generation quality.

\bibliographystyle{ACM-Reference-Format}
\bibliography{diffgar}

%%% -*-BibTeX-*-
%%% Do NOT edit. File created by BibTeX with style
%%% ACM-Reference-Format-Journals [18-Jan-2012].

\begin{thebibliography}{47}

%%% ====================================================================
%%% NOTE TO THE USER: you can override these defaults by providing
%%% customized versions of any of these macros before the \bibliography
%%% command.  Each of them MUST provide its own final punctuation,
%%% except for \shownote{}, \showDOI{}, and \showURL{}.  The latter two
%%% do not use final punctuation, in order to avoid confusing it with
%%% the Web address.
%%%
%%% To suppress output of a particular field, define its macro to expand
%%% to an empty string, or better, \unskip, like this:
%%%
%%% \newcommand{\showDOI}[1]{\unskip}   % LaTeX syntax
%%%
%%% \def \showDOI #1{\unskip}           % plain TeX syntax
%%%
%%% ====================================================================

\ifx \showCODEN    \undefined \def \showCODEN     #1{\unskip}     \fi
\ifx \showDOI      \undefined \def \showDOI       #1{#1}\fi
\ifx \showISBNx    \undefined \def \showISBNx     #1{\unskip}     \fi
\ifx \showISBNxiii \undefined \def \showISBNxiii  #1{\unskip}     \fi
\ifx \showISSN     \undefined \def \showISSN      #1{\unskip}     \fi
\ifx \showLCCN     \undefined \def \showLCCN      #1{\unskip}     \fi
\ifx \shownote     \undefined \def \shownote      #1{#1}          \fi
\ifx \showarticletitle \undefined \def \showarticletitle #1{#1}   \fi
\ifx \showURL      \undefined \def \showURL       {\relax}        \fi
% The following commands are used for tagged output and should be
% invisible to TeX
\providecommand\bibfield[2]{#2}
\providecommand\bibinfo[2]{#2}
\providecommand\natexlab[1]{#1}
\providecommand\showeprint[2][]{arXiv:#2}

\bibitem[Behrmann et~al\mbox{.}(2019)]%
        {behrmann2019invertible}
\bibfield{author}{\bibinfo{person}{Jens Behrmann}, \bibinfo{person}{Will
  Grathwohl}, \bibinfo{person}{Ricky~TQ Chen}, \bibinfo{person}{David
  Duvenaud}, {and} \bibinfo{person}{J{\"o}rn-Henrik Jacobsen}.}
  \bibinfo{year}{2019}\natexlab{}.
\newblock \showarticletitle{Invertible residual networks}.
\newblock \bibinfo{journal}{\emph{ICML}}.
\newblock


\bibitem[Chen et~al\mbox{.}(2021)]%
        {chen2020wavegrad}
\bibfield{author}{\bibinfo{person}{Nanxin Chen}, \bibinfo{person}{Yu Zhang},
  \bibinfo{person}{Heiga Zen}, \bibinfo{person}{Ron~J Weiss},
  \bibinfo{person}{Mohammad Norouzi}, {and} \bibinfo{person}{William Chan}.}
  \bibinfo{year}{2021}\natexlab{}.
\newblock \showarticletitle{WaveGrad: Estimating gradients for waveform
  generation}.
\newblock \bibinfo{journal}{\emph{ICLR}} (\bibinfo{year}{2021}).
\newblock


\bibitem[Choi et~al\mbox{.}(2022)]%
        {choi2022perception}
\bibfield{author}{\bibinfo{person}{Jooyoung Choi}, \bibinfo{person}{Jungbeom
  Lee}, \bibinfo{person}{Chaehun Shin}, \bibinfo{person}{Sungwon Kim},
  \bibinfo{person}{Hyunwoo Kim}, {and} \bibinfo{person}{Sungroh Yoon}.}
  \bibinfo{year}{2022}\natexlab{}.
\newblock \showarticletitle{Perception Prioritized Training of Diffusion
  Models}.
\newblock \bibinfo{journal}{\emph{CVPR}}.
\newblock


\bibitem[Choi et~al\mbox{.}(2020)]%
        {choi2020stargan}
\bibfield{author}{\bibinfo{person}{Yunjey Choi}, \bibinfo{person}{Youngjung
  Uh}, \bibinfo{person}{Jaejun Yoo}, {and} \bibinfo{person}{Jung-Woo Ha}.}
  \bibinfo{year}{2020}\natexlab{}.
\newblock \showarticletitle{Stargan v2: Diverse image synthesis for multiple
  domains}.
\newblock \bibinfo{journal}{\emph{CVPR}}.
\newblock


\bibitem[Devlin et~al\mbox{.}(2018)]%
        {devlin2018bert}
\bibfield{author}{\bibinfo{person}{Jacob Devlin}, \bibinfo{person}{Ming-Wei
  Chang}, \bibinfo{person}{Kenton Lee}, {and} \bibinfo{person}{Kristina
  Toutanova}.} \bibinfo{year}{2018}\natexlab{}.
\newblock \showarticletitle{Bert: Pre-training of deep bidirectional
  transformers for language understanding}.
\newblock \bibinfo{journal}{\emph{arXiv preprint arXiv:1810.04805}}
  (\bibinfo{year}{2018}).
\newblock


\bibitem[Dhariwal and Nichol(2021)]%
        {Dhariwal2021DiffusionMB}
\bibfield{author}{\bibinfo{person}{Prafulla Dhariwal} {and}
  \bibinfo{person}{Alex Nichol}.} \bibinfo{year}{2021}\natexlab{}.
\newblock \showarticletitle{Diffusion Models Beat GANs on Image Synthesis}.
\newblock \bibinfo{journal}{\emph{NeurIPS}} (\bibinfo{year}{2021}).
\newblock


\bibitem[El~Helou and S{\"u}sstrunk(2020)]%
        {el2020blind}
\bibfield{author}{\bibinfo{person}{Majed El~Helou} {and}
  \bibinfo{person}{Sabine S{\"u}sstrunk}.} \bibinfo{year}{2020}\natexlab{}.
\newblock \showarticletitle{Blind universal Bayesian image denoising with
  Gaussian noise level learning}.
\newblock \bibinfo{journal}{\emph{TIP}}  \bibinfo{volume}{29}
  (\bibinfo{year}{2020}).
\newblock


\bibitem[Esser et~al\mbox{.}(2021)]%
        {esser2021taming}
\bibfield{author}{\bibinfo{person}{Patrick Esser}, \bibinfo{person}{Robin
  Rombach}, {and} \bibinfo{person}{Bjorn Ommer}.}
  \bibinfo{year}{2021}\natexlab{}.
\newblock \showarticletitle{Taming transformers for high-resolution image
  synthesis}.
\newblock \bibinfo{journal}{\emph{CVPR}}.
\newblock


\bibitem[Goodfellow et~al\mbox{.}(2014)]%
        {goodfellow2014generative}
\bibfield{author}{\bibinfo{person}{Ian Goodfellow}, \bibinfo{person}{Jean
  Pouget-Abadie}, \bibinfo{person}{Mehdi Mirza}, \bibinfo{person}{Bing Xu},
  \bibinfo{person}{David Warde-Farley}, \bibinfo{person}{Sherjil Ozair},
  \bibinfo{person}{Aaron Courville}, {and} \bibinfo{person}{Yoshua Bengio}.}
  \bibinfo{year}{2014}\natexlab{}.
\newblock \showarticletitle{Generative adversarial nets}.
\newblock \bibinfo{journal}{\emph{NeurIPS}}  \bibinfo{volume}{27}
  (\bibinfo{year}{2014}).
\newblock


\bibitem[Heusel et~al\mbox{.}(2017)]%
        {heusel2017gans}
\bibfield{author}{\bibinfo{person}{Martin Heusel}, \bibinfo{person}{Hubert
  Ramsauer}, \bibinfo{person}{Thomas Unterthiner}, \bibinfo{person}{Bernhard
  Nessler}, {and} \bibinfo{person}{Sepp Hochreiter}.}
  \bibinfo{year}{2017}\natexlab{}.
\newblock \showarticletitle{Gans trained by a two time-scale update rule
  converge to a local nash equilibrium}.
\newblock \bibinfo{journal}{\emph{NeurIPS}} (\bibinfo{year}{2017}).
\newblock


\bibitem[Ho et~al\mbox{.}(2020)]%
        {ho2020denoising}
\bibfield{author}{\bibinfo{person}{Jonathan Ho}, \bibinfo{person}{Ajay Jain},
  {and} \bibinfo{person}{Pieter Abbeel}.} \bibinfo{year}{2020}\natexlab{}.
\newblock \showarticletitle{Denoising diffusion probabilistic models}.
\newblock \bibinfo{journal}{\emph{NeurIPS}}  \bibinfo{volume}{33}
  (\bibinfo{year}{2020}).
\newblock


\bibitem[Ho and Salimans(2021)]%
        {ho2021classifier}
\bibfield{author}{\bibinfo{person}{Jonathan Ho} {and} \bibinfo{person}{Tim
  Salimans}.} \bibinfo{year}{2021}\natexlab{}.
\newblock \showarticletitle{Classifier-free diffusion guidance}.
\newblock \bibinfo{journal}{\emph{NeurIPS 2021 Workshop on Deep Generative
  Models and Downstream Applications}}.
\newblock


\bibitem[Issenhuth et~al\mbox{.}(2021)]%
        {Issenhuth2021EdiBERTAG}
\bibfield{author}{\bibinfo{person}{Thibaut Issenhuth}, \bibinfo{person}{Ugo
  Tanielian}, \bibinfo{person}{J{\'e}r{\'e}mie Mary}, {and}
  \bibinfo{person}{David Picard}.} \bibinfo{year}{2021}\natexlab{}.
\newblock \showarticletitle{EdiBERT, a generative model for image editing}.
\newblock \bibinfo{journal}{\emph{ArXiv}}  \bibinfo{volume}{abs/2111.15264}
  (\bibinfo{year}{2021}).
\newblock


\bibitem[Jeong et~al\mbox{.}(2022)]%
        {jeong2022unsupervised}
\bibfield{author}{\bibinfo{person}{Haedong Jeong}, \bibinfo{person}{Jiyeon
  Han}, {and} \bibinfo{person}{Jaesik Choi}.} \bibinfo{year}{2022}\natexlab{}.
\newblock \showarticletitle{An Unsupervised Way to Understand Artifact
  Generating Internal Units in Generative Neural Networks}.
\newblock \bibinfo{journal}{\emph{AAAI}}.
\newblock


\bibitem[Kadkhodaie and Simoncelli(2020)]%
        {kadkhodaie2020solving}
\bibfield{author}{\bibinfo{person}{Zahra Kadkhodaie} {and}
  \bibinfo{person}{Eero~P Simoncelli}.} \bibinfo{year}{2020}\natexlab{}.
\newblock \showarticletitle{Solving linear inverse problems using the prior
  implicit in a denoiser}.
\newblock \bibinfo{journal}{\emph{arXiv preprint arXiv:2007.13640}}
  (\bibinfo{year}{2020}).
\newblock


\bibitem[Karras et~al\mbox{.}(2019)]%
        {karras2019style}
\bibfield{author}{\bibinfo{person}{Tero Karras}, \bibinfo{person}{Samuli
  Laine}, {and} \bibinfo{person}{Timo Aila}.} \bibinfo{year}{2019}\natexlab{}.
\newblock \showarticletitle{A style-based generator architecture for generative
  adversarial networks}.
\newblock \bibinfo{journal}{\emph{CVPR}}.
\newblock


\bibitem[Karras et~al\mbox{.}(2020)]%
        {Karras2019stylegan2}
\bibfield{author}{\bibinfo{person}{Tero Karras}, \bibinfo{person}{Samuli
  Laine}, \bibinfo{person}{Miika Aittala}, \bibinfo{person}{Janne Hellsten},
  \bibinfo{person}{Jaakko Lehtinen}, {and} \bibinfo{person}{Timo Aila}.}
  \bibinfo{year}{2020}\natexlab{}.
\newblock \showarticletitle{Analyzing and Improving the Image Quality of
  {StyleGAN}}.
\newblock \bibinfo{journal}{\emph{CVPR}}.
\newblock


\bibitem[Kawar et~al\mbox{.}(2022)]%
        {kawar2022denoising}
\bibfield{author}{\bibinfo{person}{Bahjat Kawar}, \bibinfo{person}{Michael
  Elad}, \bibinfo{person}{Stefano Ermon}, {and} \bibinfo{person}{Jiaming
  Song}.} \bibinfo{year}{2022}\natexlab{}.
\newblock \showarticletitle{Denoising Diffusion Restoration Models}.
\newblock \bibinfo{journal}{\emph{arXiv preprint arXiv:2201.11793}}
  (\bibinfo{year}{2022}).
\newblock


\bibitem[Kim and Ye(2022)]%
        {kim2021diffusionclip}
\bibfield{author}{\bibinfo{person}{Gwanghyun Kim} {and}
  \bibinfo{person}{Jong~Chul Ye}.} \bibinfo{year}{2022}\natexlab{}.
\newblock \showarticletitle{Diffusionclip: Text-guided image manipulation using
  diffusion models}.
\newblock \bibinfo{journal}{\emph{CVPR}} (\bibinfo{year}{2022}).
\newblock


\bibitem[Kingma and Dhariwal(2018)]%
        {kingma2018glow}
\bibfield{author}{\bibinfo{person}{Durk~P Kingma} {and}
  \bibinfo{person}{Prafulla Dhariwal}.} \bibinfo{year}{2018}\natexlab{}.
\newblock \showarticletitle{Glow: Generative flow with invertible 1x1
  convolutions}.
\newblock \bibinfo{journal}{\emph{NeurIPS}} (\bibinfo{year}{2018}).
\newblock


\bibitem[Kingma and Welling(2014)]%
        {Kingma2014AutoEncodingVB}
\bibfield{author}{\bibinfo{person}{Diederik~P. Kingma} {and}
  \bibinfo{person}{Max Welling}.} \bibinfo{year}{2014}\natexlab{}.
\newblock \showarticletitle{Auto-Encoding Variational Bayes}.
\newblock \bibinfo{journal}{\emph{ICLR}} (\bibinfo{year}{2014}).
\newblock


\bibitem[Kong et~al\mbox{.}(2020)]%
        {kong2020diffwave}
\bibfield{author}{\bibinfo{person}{Zhifeng Kong}, \bibinfo{person}{Wei Ping},
  \bibinfo{person}{Jiaji Huang}, \bibinfo{person}{Kexin Zhao}, {and}
  \bibinfo{person}{Bryan Catanzaro}.} \bibinfo{year}{2020}\natexlab{}.
\newblock \showarticletitle{Diffwave: A versatile diffusion model for audio
  synthesis}.
\newblock \bibinfo{journal}{\emph{arXiv preprint arXiv:2009.09761}}
  (\bibinfo{year}{2020}).
\newblock


\bibitem[Kuznetsova et~al\mbox{.}(2020)]%
        {kuznetsova2020open}
\bibfield{author}{\bibinfo{person}{Alina Kuznetsova}, \bibinfo{person}{Hassan
  Rom}, \bibinfo{person}{Neil Alldrin}, \bibinfo{person}{Jasper Uijlings},
  \bibinfo{person}{Ivan Krasin}, \bibinfo{person}{Jordi Pont-Tuset},
  \bibinfo{person}{Shahab Kamali}, \bibinfo{person}{Stefan Popov},
  \bibinfo{person}{Matteo Malloci}, \bibinfo{person}{Alexander Kolesnikov},
  {et~al\mbox{.}}} \bibinfo{year}{2020}\natexlab{}.
\newblock \showarticletitle{The open images dataset v4}.
\newblock \bibinfo{journal}{\emph{IJCV}} (\bibinfo{year}{2020}).
\newblock


\bibitem[Meng et~al\mbox{.}(2021)]%
        {meng2021sdedit}
\bibfield{author}{\bibinfo{person}{Chenlin Meng}, \bibinfo{person}{Yang Song},
  \bibinfo{person}{Jiaming Song}, \bibinfo{person}{Jiajun Wu},
  \bibinfo{person}{Jun-Yan Zhu}, {and} \bibinfo{person}{Stefano Ermon}.}
  \bibinfo{year}{2021}\natexlab{}.
\newblock \showarticletitle{Sdedit: Image synthesis and editing with stochastic
  differential equations}.
\newblock \bibinfo{journal}{\emph{arXiv preprint arXiv:2108.01073}}
  (\bibinfo{year}{2021}).
\newblock


\bibitem[Nichol et~al\mbox{.}(2021)]%
        {nichol2021glide}
\bibfield{author}{\bibinfo{person}{Alex Nichol}, \bibinfo{person}{Prafulla
  Dhariwal}, \bibinfo{person}{Aditya Ramesh}, \bibinfo{person}{Pranav Shyam},
  \bibinfo{person}{Pamela Mishkin}, \bibinfo{person}{Bob McGrew},
  \bibinfo{person}{Ilya Sutskever}, {and} \bibinfo{person}{Mark Chen}.}
  \bibinfo{year}{2021}\natexlab{}.
\newblock \showarticletitle{Glide: Towards photorealistic image generation and
  editing with text-guided diffusion models}.
\newblock \bibinfo{journal}{\emph{arXiv preprint arXiv:2112.10741}}
  (\bibinfo{year}{2021}).
\newblock


\bibitem[Ostrovski et~al\mbox{.}(2018)]%
        {ostrovski2018autoregressive}
\bibfield{author}{\bibinfo{person}{Georg Ostrovski}, \bibinfo{person}{Will
  Dabney}, {and} \bibinfo{person}{R{\'e}mi Munos}.}
  \bibinfo{year}{2018}\natexlab{}.
\newblock \showarticletitle{Autoregressive quantile networks for generative
  modeling}.
\newblock \bibinfo{journal}{\emph{ICML}}.
\newblock


\bibitem[Parmar et~al\mbox{.}(2018)]%
        {Parmar2018ImageT}
\bibfield{author}{\bibinfo{person}{Niki Parmar}, \bibinfo{person}{Ashish
  Vaswani}, \bibinfo{person}{Jakob Uszkoreit}, \bibinfo{person}{Lukasz Kaiser},
  \bibinfo{person}{Noam~M. Shazeer}, \bibinfo{person}{Alexander Ku}, {and}
  \bibinfo{person}{Dustin Tran}.} \bibinfo{year}{2018}\natexlab{}.
\newblock \showarticletitle{Image Transformer}.
\newblock \bibinfo{journal}{\emph{ICML}}.
\newblock


\bibitem[Paszke et~al\mbox{.}(2019)]%
        {paszke2019pytorch}
\bibfield{author}{\bibinfo{person}{Adam Paszke}, \bibinfo{person}{Sam Gross},
  \bibinfo{person}{Francisco Massa}, \bibinfo{person}{Adam Lerer},
  \bibinfo{person}{James Bradbury}, \bibinfo{person}{Gregory Chanan},
  \bibinfo{person}{Trevor Killeen}, \bibinfo{person}{Zeming Lin},
  \bibinfo{person}{Natalia Gimelshein}, \bibinfo{person}{Luca Antiga},
  {et~al\mbox{.}}} \bibinfo{year}{2019}\natexlab{}.
\newblock \showarticletitle{Pytorch: An imperative style, high-performance deep
  learning library}.
\newblock \bibinfo{journal}{\emph{NeurIPS}} (\bibinfo{year}{2019}).
\newblock


\bibitem[Radford et~al\mbox{.}(2021)]%
        {radford2021learning}
\bibfield{author}{\bibinfo{person}{Alec Radford}, \bibinfo{person}{Jong~Wook
  Kim}, \bibinfo{person}{Chris Hallacy}, \bibinfo{person}{Aditya Ramesh},
  \bibinfo{person}{Gabriel Goh}, \bibinfo{person}{Sandhini Agarwal},
  \bibinfo{person}{Girish Sastry}, \bibinfo{person}{Amanda Askell},
  \bibinfo{person}{Pamela Mishkin}, \bibinfo{person}{Jack Clark},
  {et~al\mbox{.}}} \bibinfo{year}{2021}\natexlab{}.
\newblock \showarticletitle{Learning transferable visual models from natural
  language supervision}.
\newblock \bibinfo{journal}{\emph{ICML}}.
\newblock


\bibitem[Ramachandran et~al\mbox{.}(2017)]%
        {ramachandran2017searching}
\bibfield{author}{\bibinfo{person}{Prajit Ramachandran},
  \bibinfo{person}{Barret Zoph}, {and} \bibinfo{person}{Quoc~V Le}.}
  \bibinfo{year}{2017}\natexlab{}.
\newblock \showarticletitle{Searching for activation functions}.
\newblock \bibinfo{journal}{\emph{arXiv preprint arXiv:1710.05941}}
  (\bibinfo{year}{2017}).
\newblock


\bibitem[Rombach et~al\mbox{.}(2022)]%
        {rombach2022high}
\bibfield{author}{\bibinfo{person}{Robin Rombach}, \bibinfo{person}{Andreas
  Blattmann}, \bibinfo{person}{Dominik Lorenz}, \bibinfo{person}{Patrick
  Esser}, {and} \bibinfo{person}{Bj{\"o}rn Ommer}.}
  \bibinfo{year}{2022}\natexlab{}.
\newblock \showarticletitle{High-resolution image synthesis with latent
  diffusion models}.
\newblock \bibinfo{journal}{\emph{CVPR}}.
\newblock


\bibitem[Ronneberger et~al\mbox{.}(2015)]%
        {ronneberger2015u}
\bibfield{author}{\bibinfo{person}{Olaf Ronneberger}, \bibinfo{person}{Philipp
  Fischer}, {and} \bibinfo{person}{Thomas Brox}.}
  \bibinfo{year}{2015}\natexlab{}.
\newblock \showarticletitle{U-net: Convolutional networks for biomedical image
  segmentation}.
\newblock \bibinfo{journal}{\emph{International Conference on Medical image
  computing and computer-assisted intervention}}, \bibinfo{pages}{234--241}.
\newblock


\bibitem[Saharia et~al\mbox{.}(2021a)]%
        {saharia2021palette}
\bibfield{author}{\bibinfo{person}{Chitwan Saharia}, \bibinfo{person}{William
  Chan}, \bibinfo{person}{Huiwen Chang}, \bibinfo{person}{Chris~A Lee},
  \bibinfo{person}{Jonathan Ho}, \bibinfo{person}{Tim Salimans},
  \bibinfo{person}{David~J Fleet}, {and} \bibinfo{person}{Mohammad Norouzi}.}
  \bibinfo{year}{2021}\natexlab{a}.
\newblock \showarticletitle{Palette: Image-to-image diffusion models}.
\newblock \bibinfo{journal}{\emph{arXiv preprint arXiv:2111.05826}}
  (\bibinfo{year}{2021}).
\newblock


\bibitem[Saharia et~al\mbox{.}(2021b)]%
        {saharia2021image}
\bibfield{author}{\bibinfo{person}{Chitwan Saharia}, \bibinfo{person}{Jonathan
  Ho}, \bibinfo{person}{William Chan}, \bibinfo{person}{Tim Salimans},
  \bibinfo{person}{David~J Fleet}, {and} \bibinfo{person}{Mohammad Norouzi}.}
  \bibinfo{year}{2021}\natexlab{b}.
\newblock \showarticletitle{Image super-resolution via iterative refinement}.
\newblock \bibinfo{journal}{\emph{arXiv preprint arXiv:2104.07636}}
  (\bibinfo{year}{2021}).
\newblock


\bibitem[Shocher et~al\mbox{.}(2018)]%
        {shocher2018zero}
\bibfield{author}{\bibinfo{person}{Assaf Shocher}, \bibinfo{person}{Nadav
  Cohen}, {and} \bibinfo{person}{Michal Irani}.}
  \bibinfo{year}{2018}\natexlab{}.
\newblock \showarticletitle{“zero-shot” super-resolution using deep
  internal learning}.
\newblock \bibinfo{journal}{\emph{CVPR}}.
\newblock


\bibitem[Sinha et~al\mbox{.}(2021)]%
        {sinha2021d2c}
\bibfield{author}{\bibinfo{person}{Abhishek Sinha}, \bibinfo{person}{Jiaming
  Song}, \bibinfo{person}{Chenlin Meng}, {and} \bibinfo{person}{Stefano
  Ermon}.} \bibinfo{year}{2021}\natexlab{}.
\newblock \showarticletitle{D2c: Diffusion-decoding models for few-shot
  conditional generation}.
\newblock \bibinfo{journal}{\emph{NeurIPS}}  \bibinfo{volume}{34}
  (\bibinfo{year}{2021}).
\newblock


\bibitem[Song et~al\mbox{.}(2021)]%
        {song2020denoising}
\bibfield{author}{\bibinfo{person}{Jiaming Song}, \bibinfo{person}{Chenlin
  Meng}, {and} \bibinfo{person}{Stefano Ermon}.}
  \bibinfo{year}{2021}\natexlab{}.
\newblock \showarticletitle{Denoising diffusion implicit models}.
\newblock \bibinfo{journal}{\emph{ICLR}} (\bibinfo{year}{2021}).
\newblock


\bibitem[Song and Ermon(2019)]%
        {Song2019GenerativeMB}
\bibfield{author}{\bibinfo{person}{Yang Song} {and} \bibinfo{person}{Stefano
  Ermon}.} \bibinfo{year}{2019}\natexlab{}.
\newblock \showarticletitle{Generative Modeling by Estimating Gradients of the
  Data Distribution}.
\newblock \bibinfo{journal}{\emph{NeurIPS}} (\bibinfo{year}{2019}).
\newblock


\bibitem[Tan et~al\mbox{.}(2021)]%
        {tan2021systematic}
\bibfield{author}{\bibinfo{person}{Way Tan}, \bibinfo{person}{Bihan Wen}, {and}
  \bibinfo{person}{Xulei Yang}.} \bibinfo{year}{2021}\natexlab{}.
\newblock \showarticletitle{Systematic analysis and removal of circular
  artifacts for stylegan}.
\newblock \bibinfo{journal}{\emph{arXiv preprint arXiv:2103.01090}}
  (\bibinfo{year}{2021}).
\newblock


\bibitem[Tousi et~al\mbox{.}(2021)]%
        {tousi2021automatic}
\bibfield{author}{\bibinfo{person}{Ali Tousi}, \bibinfo{person}{Haedong Jeong},
  \bibinfo{person}{Jiyeon Han}, \bibinfo{person}{Hwanil Choi}, {and}
  \bibinfo{person}{Jaesik Choi}.} \bibinfo{year}{2021}\natexlab{}.
\newblock \showarticletitle{Automatic correction of internal units in
  generative neural networks}.
\newblock \bibinfo{journal}{\emph{CVPR}}.
\newblock


\bibitem[Vahdat and Kautz(2020)]%
        {Vahdat2020NVAEAD}
\bibfield{author}{\bibinfo{person}{Arash Vahdat} {and} \bibinfo{person}{Jan
  Kautz}.} \bibinfo{year}{2020}\natexlab{}.
\newblock \showarticletitle{NVAE: A Deep Hierarchical Variational Autoencoder}.
\newblock \bibinfo{journal}{\emph{NeurIPS}} (\bibinfo{year}{2020}).
\newblock


\bibitem[Van Den~Oord et~al\mbox{.}(2017)]%
        {van2017neural}
\bibfield{author}{\bibinfo{person}{Aaron Van Den~Oord}, \bibinfo{person}{Oriol
  Vinyals}, {et~al\mbox{.}}} \bibinfo{year}{2017}\natexlab{}.
\newblock \showarticletitle{Neural discrete representation learning}.
\newblock \bibinfo{journal}{\emph{NeurIPS}} (\bibinfo{year}{2017}).
\newblock


\bibitem[Vaswani et~al\mbox{.}(2017)]%
        {vaswani2017attention}
\bibfield{author}{\bibinfo{person}{Ashish Vaswani}, \bibinfo{person}{Noam
  Shazeer}, \bibinfo{person}{Niki Parmar}, \bibinfo{person}{Jakob Uszkoreit},
  \bibinfo{person}{Llion Jones}, \bibinfo{person}{Aidan~N Gomez},
  \bibinfo{person}{{\L}ukasz Kaiser}, {and} \bibinfo{person}{Illia
  Polosukhin}.} \bibinfo{year}{2017}\natexlab{}.
\newblock \showarticletitle{Attention is all you need}.
\newblock \bibinfo{journal}{\emph{NeurIPS}} (\bibinfo{year}{2017}).
\newblock


\bibitem[Wang et~al\mbox{.}(2021a)]%
        {wang2021towards}
\bibfield{author}{\bibinfo{person}{Xintao Wang}, \bibinfo{person}{Yu Li},
  \bibinfo{person}{Honglun Zhang}, {and} \bibinfo{person}{Ying Shan}.}
  \bibinfo{year}{2021}\natexlab{a}.
\newblock \showarticletitle{Towards real-world blind face restoration with
  generative facial prior}.
\newblock \bibinfo{journal}{\emph{CVPR}}.
\newblock


\bibitem[Wang et~al\mbox{.}(2021b)]%
        {wang2021real}
\bibfield{author}{\bibinfo{person}{Xintao Wang}, \bibinfo{person}{Liangbin
  Xie}, \bibinfo{person}{Chao Dong}, {and} \bibinfo{person}{Ying Shan}.}
  \bibinfo{year}{2021}\natexlab{b}.
\newblock \showarticletitle{Real-esrgan: Training real-world blind
  super-resolution with pure synthetic data}. In
  \bibinfo{booktitle}{\emph{ICCV}}.
\newblock


\bibitem[Wu and He(2018)]%
        {wu2018group}
\bibfield{author}{\bibinfo{person}{Yuxin Wu} {and} \bibinfo{person}{Kaiming
  He}.} \bibinfo{year}{2018}\natexlab{}.
\newblock \showarticletitle{Group normalization}. In
  \bibinfo{booktitle}{\emph{ECCV}}.
\newblock


\bibitem[Zhang et~al\mbox{.}(2019)]%
        {zhang2019detecting}
\bibfield{author}{\bibinfo{person}{Xu Zhang}, \bibinfo{person}{Svebor Karaman},
  {and} \bibinfo{person}{Shih-Fu Chang}.} \bibinfo{year}{2019}\natexlab{}.
\newblock \showarticletitle{Detecting and simulating artifacts in gan fake
  images}.
\newblock \bibinfo{journal}{\emph{WIFS}}.
\newblock


\end{thebibliography}
\clearpage

\setcounter{section}{0}
\renewcommand\thesection{\Alph{section}}

\section{Training Details}
To demonstrate the capabilities of DiffGAR, we conduct experiments on the proposed FFHQ and AFHQ-Dog artifact datasets.
All our quantitative and qualitative results in the main paper are $256$ image resolution.
We use the standard Adam optimizer with a fixed $1e-4$ learning rate.
We set the total timestep $T$ to $1000$.
We implemented our model DiffGAR with PyTorch~\cite{paszke2019pytorch} and train them on $8$ GPUs for $3$ days around.

\section{Test Dataset Construction}
For FFHQ dataset, we generate 2k images for inference. For AFHQ-Dog dataset, 500 images are used for inference.
We use the code from ~\cite{esser2021taming} to simulate GPT sampling artifacts, for FFHQ dataset, we set the number of top-k to 500 and the sampling temperature to 18. 
For AFHQ-Dog dataset, we set the number of top-k to 600 and the sampling temperature to 22. 
It is worth noting that the same as the training phase, in order to keep the identity of the object in the image, we set $90\%$ of image tokens unchanged and only $10\%$ of the tokens will be resampled using the given temperature and top-k. 
For replace\_token artifacts, we randomly sample a $4 \times 4$ rectangle and replace the tokens within the selected rectangle with randomly sampled tokens from the codebook.
For ddim\_gaussian artifacts, we set $T_0$ to $840$, forward steps and inversion steps to $48$, $\alpha$ to 0.25. 
Ddim\_scale artifacts share the same diffusion steps as ddim\_gaussian, and $\gamma$, $\beta$ is set to be $1.009$, $0.005$ for FFHQ dataset.
For AFHQ-Dog, we set $\gamma$ to $1.015$, $\beta$ to $0.001$.

\section{Model Architectures}
In this section, we will present the details of the models used in the first and second stages of DiffGAR, respectively.

\subsection{Details on Autoencoder Models}
We use the pretrained autoencoder model trained on OpenImages~\cite{kuznetsova2020open} in VQ-regularized latent spaces from Latent Diffusion.
The architecture of the convolutional encoder and decoder models used in autoencoder  is shown in Tab.~\ref{tab:autoencoder_high_level}.
The Non-Local Block means the self-attention module.
We set the downsampling steps $m$ to $2$. The codebook size $K$ is $8192$ and the dimensionality of codebook entries is $3$.
When taking an artifact or original image as input, the autoencoder can map it into a latent representation $z \in \mathbb{R}^{3  \times 64 \times 64 }$.

\begin{table*}
\caption{\textbf{High-level architecture of the encoder and decoder of the autoencoder.}}
  \begin{center}
    {\scalebox{0.95}{
\begin{tabular}{c|c  }
\toprule
      \textbf{Encoder} &  \textbf{Decoder}   \\
    \midrule
    $x \in  \mathbb{R}^{3  \times 256 \times 256 } $  &  $z \in \mathbb{R}^{3  \times 64 \times 64 } $ \\
    $\text{Conv2D} \to  \mathbb{R}^{128  \times 256 \times 256 } $  &  $\text{Conv2D} \to  \mathbb{R}^{3  \times 64 \times 64 } $  \\
    $ 2 \times \{\text{Residual Block, Downsample Block }\}  \to \mathbb{R}^{512  \times 64 \times 64 }  $  &  $\text{Residual Block} \to  \mathbb{R}^{512  \times 64 \times 64 } $  \\
    $\text{Residual Block} \to  \mathbb{R}^{512  \times 64 \times 64 } $  &  $\text{Non-Local Block} \to  \mathbb{R}^{512  \times 64 \times 64 } $   \\
    $\text{Non-Local Block} \to  \mathbb{R}^{512  \times 64 \times 64 } $  &  $\text{Residual Block} \to  \mathbb{R}^{512  \times 64 \times 64 } $ \\
    $\text{Residual Block} \to  \mathbb{R}^{512  \times 64 \times 64 } $  &  $ 2 \times \{\text{Residual Block, Upsample Block }\}  \to \mathbb{R}^{128  \times 256 \times 256 }  $ \\
    $\text{GroupNorm, Swish, Conv2D} \to  \mathbb{R}^{3  \times 64 \times 64 } $  &  $\text{GroupNorm, Swish, Conv2D} \to  \mathbb{R}^{3  \times 256 \times 256 } $  \\
    \bottomrule
\end{tabular}
}}
\end{center}
\label{tab:autoencoder_high_level}
\end{table*}

\begin{figure*}
  \begin{center}
  \includegraphics[width=0.8 \linewidth]{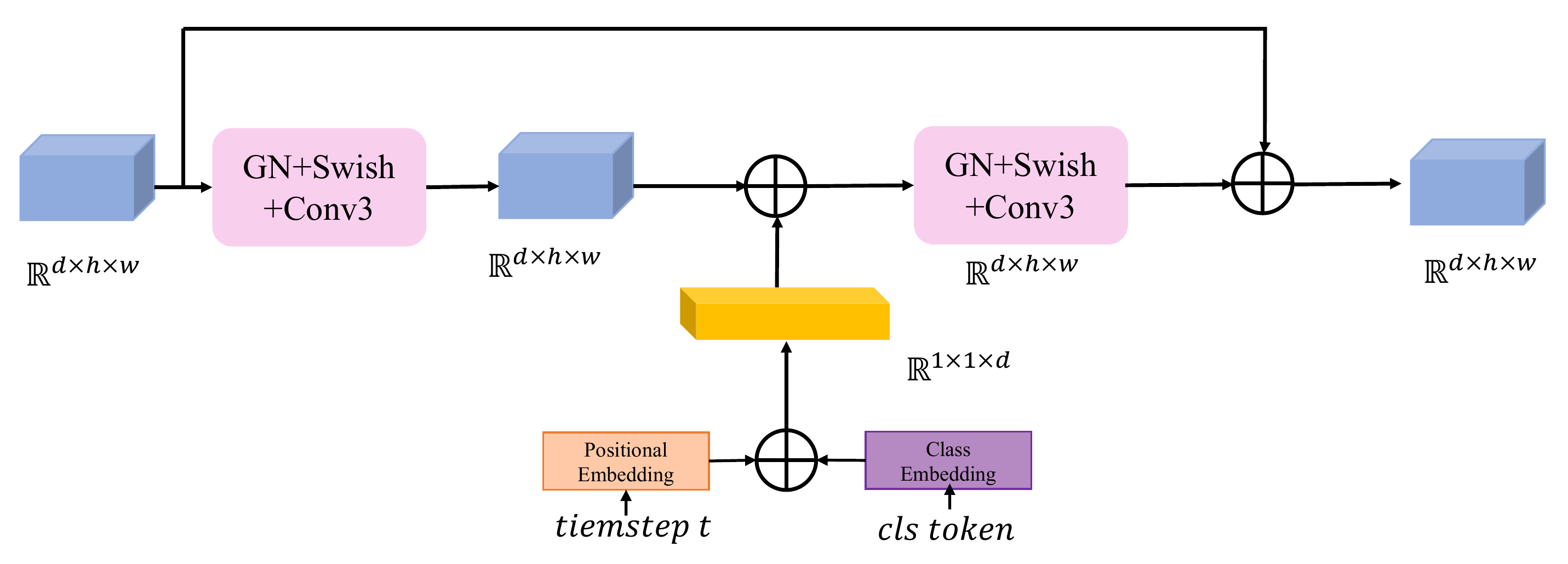}
  \end{center}
  \caption{\textbf{Details of ResBlock.}}
  \vspace{-0.3cm}
  \label{fig:resblock_detail}
\end{figure*}

\subsection{Details on Conditional Diffusion Models}

The conditional diffusion model consists of an encoder module, a middle module, a decoder module and an embedding module.
The encoder part is composed of one input convolution layer and $4$ ResBlock module, mapping the input representation $z \in \mathbb{R}^{3  \times 64 \times 64 } $ into a feature map $z_e \in \mathbb{R}^{960  \times 8 \times 8 } $.
Each ResBlock consists of two convolutional blocks including Group normalization~\cite{wu2018group} and a Swish activation module~\cite{ramachandran2017searching} with the residual connection as shown in Fig.~\ref{fig:resblock_detail}.

The middle part consists of $2$ ResBlocks and the second block is a Transformer Block.
In the decoder part, the module is composed of $4$ ResBlock module and one output convolution layer.
The intermediate feature map of the decoder whose resolution is the same as the feature map of the encoder part will be concated through skip connection. 
In the time embedding module, there are two kinds of embedding vectors.
First, we employ sinusoidal position embeddings~\cite{vaswani2017attention} to encode the diffusion timestep $t$ into a $768$-dimensional vector, so that the neural network can know at which particular time step it is operating.
In addition, we learn an artifact class embedding matrix $E \in \mathbb{R}^{7*768}$ which represents $6$ generative artifacts classes and a mask token for unknown artifact class, $768$ is the dimensionality of artifact class embedding.
Then, these two embeddings will be added together and fed to each ResBlock.
Notably, the self-attention blocks are added to the ResBlock at the resolution of $8 \times 8, 16 \times 16, 32 \times 32$ resolution.

\section{Additional Qualitative Results}
\begin{figure*}[!h]
    \centering
    \begin{subfigure}{0.45 \linewidth}
      \includegraphics[width=\linewidth%height=9cm, width=9cm
      ]{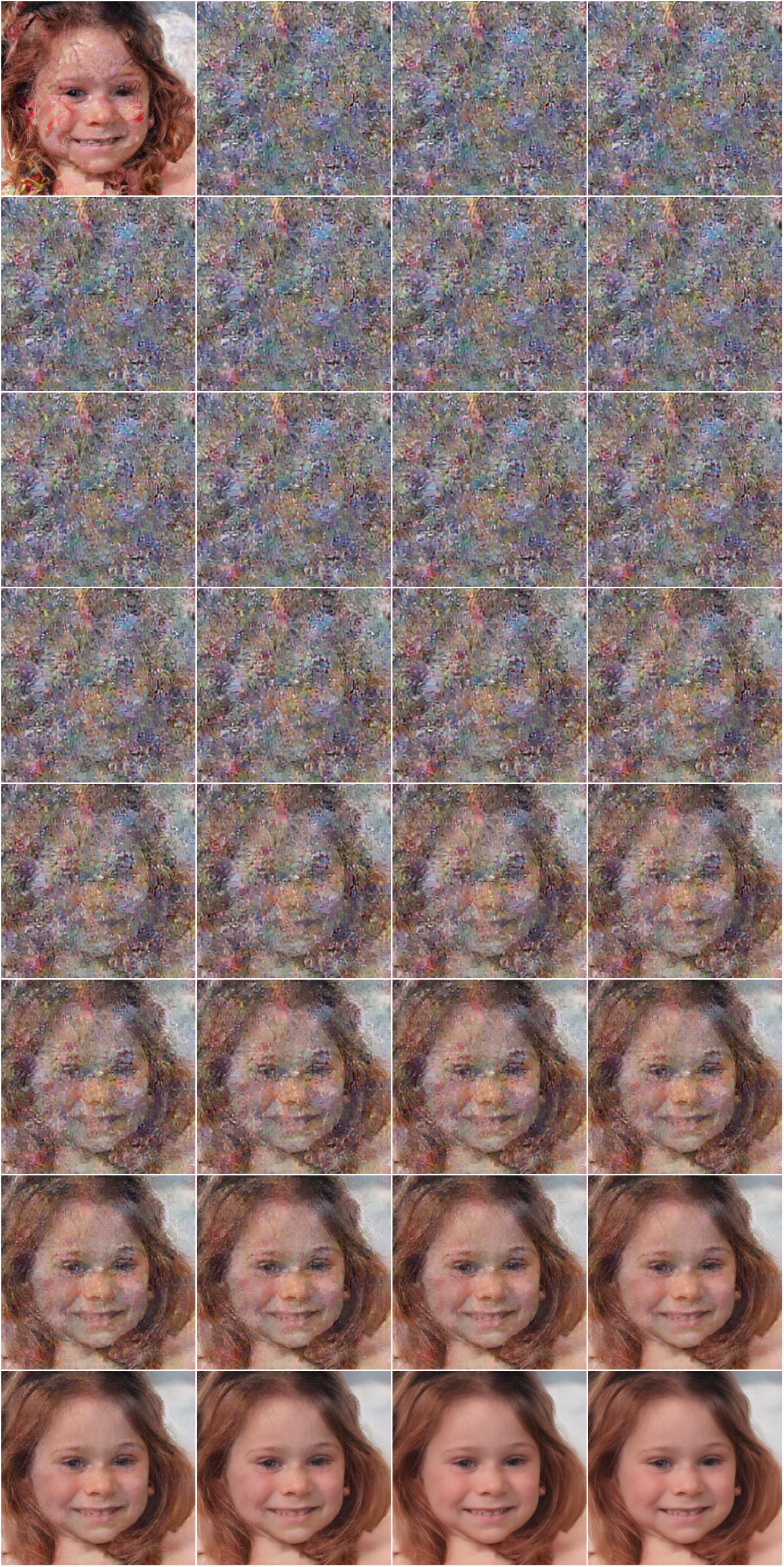} 
      \caption{Sample1.}
      \label{fig:ddim_sample1}
    \end{subfigure}
    \hfill
    \begin{subfigure}{0.45 \linewidth}
      \includegraphics[width=\linewidth %height=9cm, width=9cm
      ]{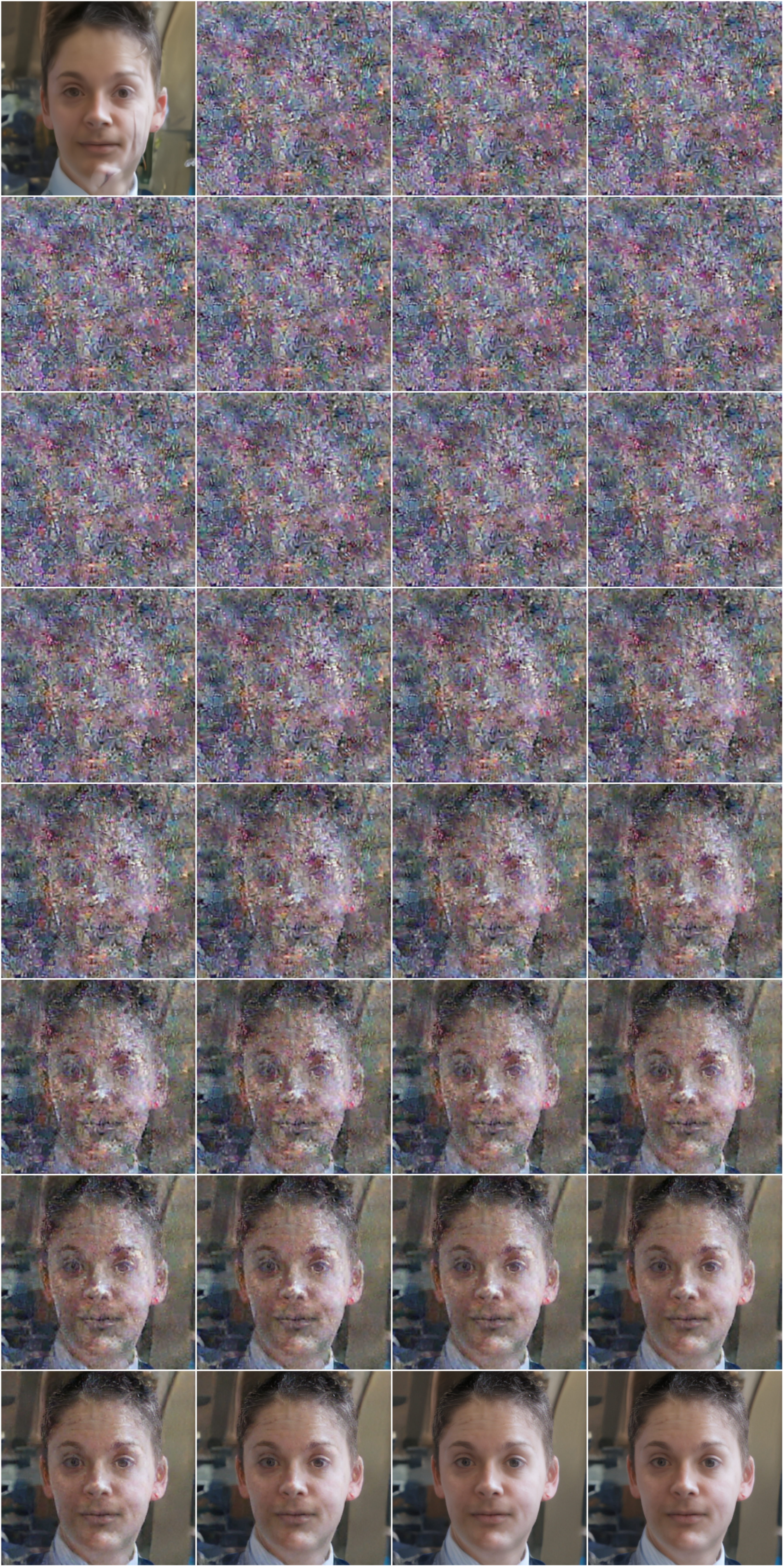}
      \caption{Sample2.}
      \label{fig:ddim_sample2}
    \end{subfigure}
    \caption{\textbf{Two samples of a completed DDIM inference process.} This figure shows the DDIM sampling process of $30$ inference steps when removing artifacts.}
    \label{fig:ddim_prcess_show}
  \end{figure*}

Finally, we provide additional qualitative results.
Fig.~\ref{fig:ddim_prcess_show} shows the DDIM sampling process of $30$ inference steps.
The first image is the sampled artifact image. 
The second image is the result after feeding the sampled Gaussian noise into the decoder of the autoencoder.
The subsequent sequential images represent the results $y_t$ generated by stepwise denoising from the sampled Gaussian noise $y_T$, as shown in line 5 of Algorithm 2 in our main paper.
From Fig.~\ref{fig:ddim_prcess_show} we can see that with the artifact image as the conditional input, DiffGAR can generate a clean image corresponding to the artifact image by gradual denoising it from the Gaussian noise.

\end{document}